%% file: main.tex
\documentclass{bmvc2k}
\usepackage{amsmath,amsfonts}
\usepackage{algorithm}
\usepackage{algorithmic}
\usepackage{graphicx}
\usepackage{textcomp}
\usepackage{bbold}
\usepackage{xcolor}
\usepackage{microtype}
\usepackage{multirow}
\usepackage{subfigure}
\usepackage{booktabs} 
\usepackage{wrapfig}
\usepackage{hyperref} 
\hypersetup{
    colorlinks=true,
    linkcolor=blue,
    citecolor = blue,
    urlcolor=blue,}
\usepackage{xspace}
\usepackage{diagbox}
\usepackage{array}
\usepackage{float}
\usepackage{tikz}
\usepackage{adjustbox}
\input{_mathdef}
\title{\Large \hspace{.2cm} Vicious Classifiers: Assessing Inference-time \\ 
\hspace{.2cm} Data Reconstruction Risk in Edge Computing
\vspace{0.7cm}
}
\addauthor{\hspace{.2cm} Mohammad Malekzadeh}{mohammad.malekzadeh@nokia.com}{1}
\addauthor{\hspace{.25cm} Deniz Gündüz}{ d.gunduz@imperial.ac.uk}{2}

\addinstitution{
 Nokia Bell Labs \\ Cambridge, UK
}
\addinstitution{
 Imperial College London \\ London, UK
}
\runninghead{Malekzadeh and Gündüz}{Assessing Data Reconstruction Risk}
\def\eg{\emph{e.g}\bmvaOneDot}

\begin{document}
\maketitle
\begin{abstract}
Privacy-preserving {\em inference}, in edge computing paradigms, encourages the \users of machine-learning services to locally run a model on their {\em private input}, and only share the {\em model's outputs} for a  {\em \pub task} with the server. We study how a {\em \vic} server can reconstruct the input data by observing only the model's outputs, while keeping the target accuracy very close to that of a {\em honest} server: by jointly training a {\em target} model (to run at \users' side) and an {\em attack} model for data reconstruction (to secretly use at \server's side). We present a new measure to assess the {\em inference-time reconstruction risk}. Evaluations on six benchmark datasets show the model's input can be approximately reconstructed from the outputs of a {\em single} inference. We propose a primary defense mechanism to distinguish {\em \vic} versus {\em \hon} classifiers at inference time. By studying such a risk associated with emerging ML services, our work has implications for enhancing privacy in edge computing. We discuss open challenges and directions for future studies and release our code as a benchmark for the community at \codelink.
\\
\end{abstract}
\section{Introduction}\label{sec_intro}
Emerging machine learning~(ML) services build profiles of their {\em \users} by collecting their {\em personal data}. \Users might share some specific data with a service provider in exchange for some {\em \pub} utility. Health monitoring, wellness recommendations, dynamic pricing, or personalized content usually attract \users to share their data. If the \users are aware of the type of data collected about them, and explicitly confirm their consent, such data collection and profiling is usually considered legitimate~\cite{gdpr2018}. However, the challenge is to ensure that the data collected by a {\em server} will only be used to deliver the \pub service they offer to their \users~\cite{kearns2019ethical}. Such data might be used to make other \priv inferences about the \user's personality or identity, which are considered data privacy attacks.

To preserve privacy, current techniques are on-device~\cite{teerapittayanon2017distributed, zhou2019edge} or encrypted~\cite{gentry2013homomorphic, bourse2018fast} computations that hide inputs, as well as all the intermediate representations computed by the {\em model}, and only release the {\em outputs} to the server. Since such edge inferences for a \pub task might not seem sensitive to \users' privacy, the model's outputs are released to the server in their raw form; as the server needs these outputs to perform their ultimate analyses and satisfy the services promised to the \users. In various situations, minimal communication between users and servers is crucial, 
for example, in tasks such as age or identity verification. The server requires the model's outputs to grant permission for the \user to proceed with subsequent actions like account creation or payment. {\em We argue that such a paradigm of running ML models at the edge and only sharing the outputs with a service provider does not guarantee a meaningful privacy protection for edge \users}. 

\begin{wrapfigure}{r}{0.53\textwidth}
    \centering
    \includegraphics[width=0.52\textwidth]{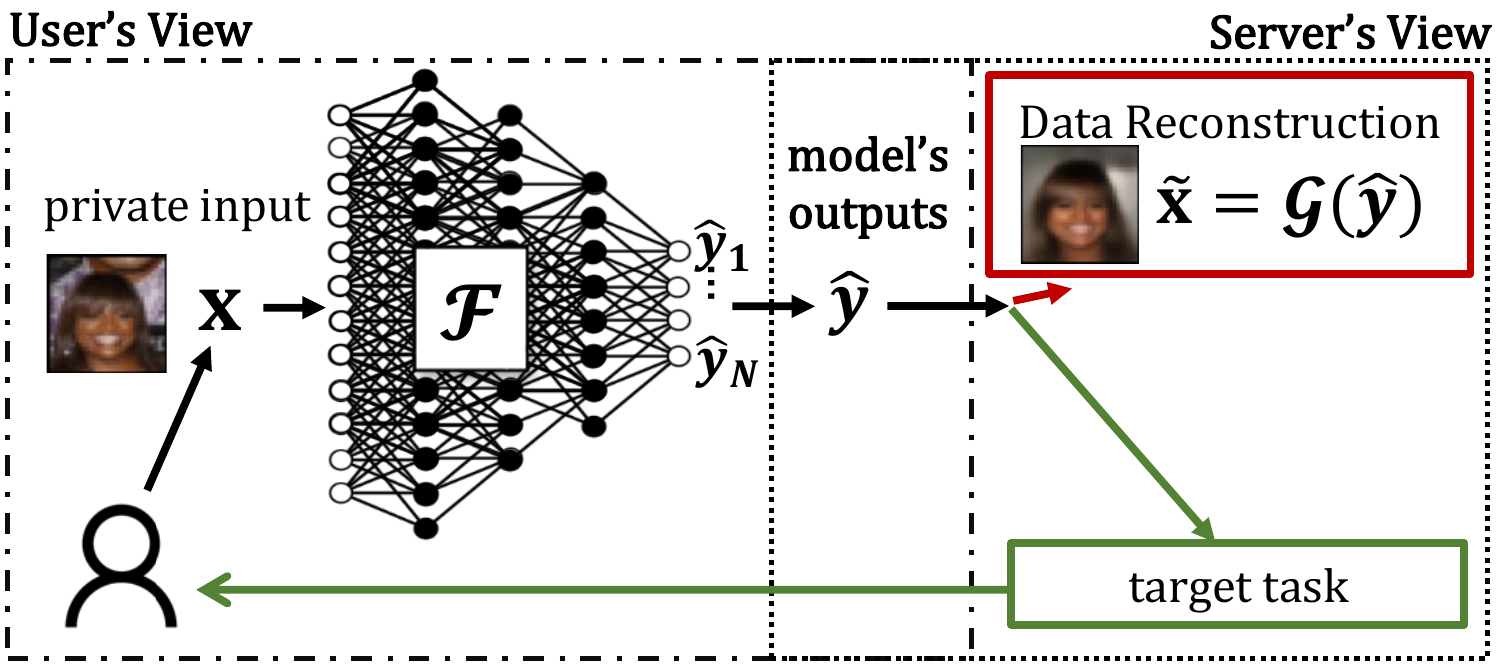}
    \caption{Processing \user's input $\bx$, the \server receives only the output $\byh\!=\!\cF(\bx)$. We show that for any model $\cF$, the \server can train an attack model $\cG$ to secretly reconstruct the input from observed output, while providing the \pub service to the \user with high accuracy.}
    \label{fig_bigpic} 
\end{wrapfigure}
As shown in Figure~\ref{fig_bigpic}, we consider a common scenario of edge or encrypted inference, in which a {\em \user} owns \priv data $\bx$, and a semi-trusted {\em \server} owns an $N$-output ML classifier $\cF$.  We put no constraint on the \user's access to $\cF$; \eg \users can have a complete white-box view of $\cF$. We assume that the \server only observes the model's output $\byh = \cF(\bx)$ (\aka$ ${\em logits}), which aims to help in predicting the \pub information $\by$. Our main assumption is that $\byh \in \bbR^{N}$ is a real-valued vector of dimension $N$, where each $\byh_n \in \bbR$, for all $n \in \{1,2, \dots, N\}$, is the {\em logit score} for the corresponding class or attribute $\by_n$.  There are several reasons that a \server might ask for observing the real-valued outputs $\byh$ to reach the ultimate decision at the \server's side; compared to only observing the $\argmax({\byh})$ or softmax($\byh$). For example, the logit scores allow the \server to perform top-K predictions, to measure the uncertainty in the estimation, or to figure out adversarial or out-of-distribution samples~\cite{malinin2018predictive}. Here are the \textbf{contributions} of our paper:

\textbf{(1)} Over-parameterized deep neural networks (DNNs) can be trained to efficiently encode additional information about their input data into the model's outputs which are supposed to reveal nothing more than a specific \pub class or attribute. We propose jointly training a multi-task model $\cF$ (\ie a {\em \vic classifiers}) as a classifier of \pub attributes as well as a decoder model $\cG$ (\ie an {\em attack model}) for reconstructing the input data from the shared outputs. The trained $\cF$ can be efficiently useful for the \pub task, and also secretly encode \priv information that allows reconstructing the \user's input data at inference time. Evaluations on MNIST, FMNIST, CIFAR10, CIFAR100, TinyImageNet, and CelebA datasets show input data can be approximately reconstructed from just the outputs of a single inference. To assess the success of a malicious \server, we consider two settings, where \users share either the {\em logits} outputs or the {\em softmax} outputs. For the same model, in the softmax setting, it is harder to establish a good trade-off between the accuracy of \pub task and the quality of reconstructed data, particularly, when the number of classes or attributes is less than  $10$.

\textbf{(2)} To measure the risk of data reconstruction, previous works~\cite{zhu2019deep, geiping2020inverting, yin2021see} mostly use mean-squared error~(MSE), peak signal-to-noise ratio~(PSNR), or structural similarity index measure~(SSIM). Euclidean distance-based measures assume that features are uncorrelated, which is not true for real-world data, such as images where pixels often have high correlation. We believe  \textit{the risk of a reconstruction attack} depends not only on the similarity of the reconstruction to the original data, but also on the likelihood of that sample data. To this end, we propose a new measure of reconstruction success rate based on the Mahalanobis distance, which considers the covariance matrix of the data. Our proposed measure, called {\em reconstruction risk}, also offers a probabilistic view on data reconstruction attack and thus, offers a principled way to evaluate the success of an attack across models and datasets.

\textbf{(3)} 
Distinguishing \hon models
from \vic ones is not trivial, and blindly applying perturbations to the outputs of all models can damage the utility received from honest servers. \Users usually observe a trained model $\cF$ that is claimed to be trained for a \pub task, but the exact training objective is unknown. Whether $\cF$ only performs the claimed task or it also secretly performs another task is unknown. To this end, we propose a method for {\em estimating the likelihood of a model being \vic}, based on the idea that a model trained only for the \pub task should not be far from the ``ideal'' solution for the \pub task; if it is only trained using the claimed objective function. On the other hand, if the model is \vic and is trained using another objective function to perform other tasks in parallel to the claimed one, then the model probably has not converged to the ideal solution for the \pub task. Our proposed defense can work even in black-box scenarios (\eg encrypted computing), and provides a practical estimation for distinguishing \hon vs. \vic models by only using a very small set of data points labeled for the \pub task (see Appendix~\S\ref{sec_defense}).
 
By uncovering a major risk in using emerging ML services, this paper helps advance privacy protection for the \users' of ML services. Our proposed analysis is just a first look, thus we conclude this paper by discussing current challenges and open directions for future investigations. We open-source our code at \codelink.  
\vspace{-0.95cm}
\section{Methodology}\label{sec_formulation}\label{sec_training}

{\bf Problem Formulation and  Threat Model.}
Let $\cX \times \cY$ denote the joint distribution over {\em data}  and {\em labels} (or {\em attributes}). We assume each data point $(\bx,\by) \sim (\cX \times \cY)$ either (1)~exclusively belongs to one of the $N$ classes (\ie categorical $\cY=\{1,2,\dots, N\}$),  or (2)~has $N$ binary attributes (\ie  $\cY=\{0,1\}^{N}$).  
Let the {\em server} train a {\em model} $\cF$ on a {\em \pub} task $\cY$, where $
\cF(\bx) = \byh = [\byh_1, \byh_2, \dots, \byh_N]$
denotes the model's {\em outputs}; \ie prediction scores (logits) over $\cY$. For the categorical case, $\cY=\{1,2,\dots, N\}$, each $\byh_i$ shows the logit score for the class $i$ (\eg the score for class $i$ in CIFAR10 dataset). For the binary case, $\cY=\{0,1\}^{N}$, each $\byh_i$ shows the logit score for the attribute $i$ (\eg the score for attribute $i$ in CelebA dataset, such as ``smiling'' attribute). We allow $\cF$ to have any arbitrary architecture; \eg to be a single model with $N$ outputs, or to be an ensemble of $N$ models each with a single output, or any other architectural choice. Model $\cF$ is trained by the server (which acts as the attacker), thus 
$\cF$ is white-box to the server. At test time, the \users will have a complete white-box view of $\cF$. We consider two settings: (1) {\em logit outputs}, where $\byh \in \bbR^{N}$, and (2) {\em softmax outputs}, where using the standard softmax function, $\byh$ is normalized to a probability distribution over the possible classes (see Appendix~\S\ref{sec_threat_model}).

\begin{wrapfigure}{r}{0.45\textwidth}
    \centering
    \includegraphics[width=.43\textwidth]{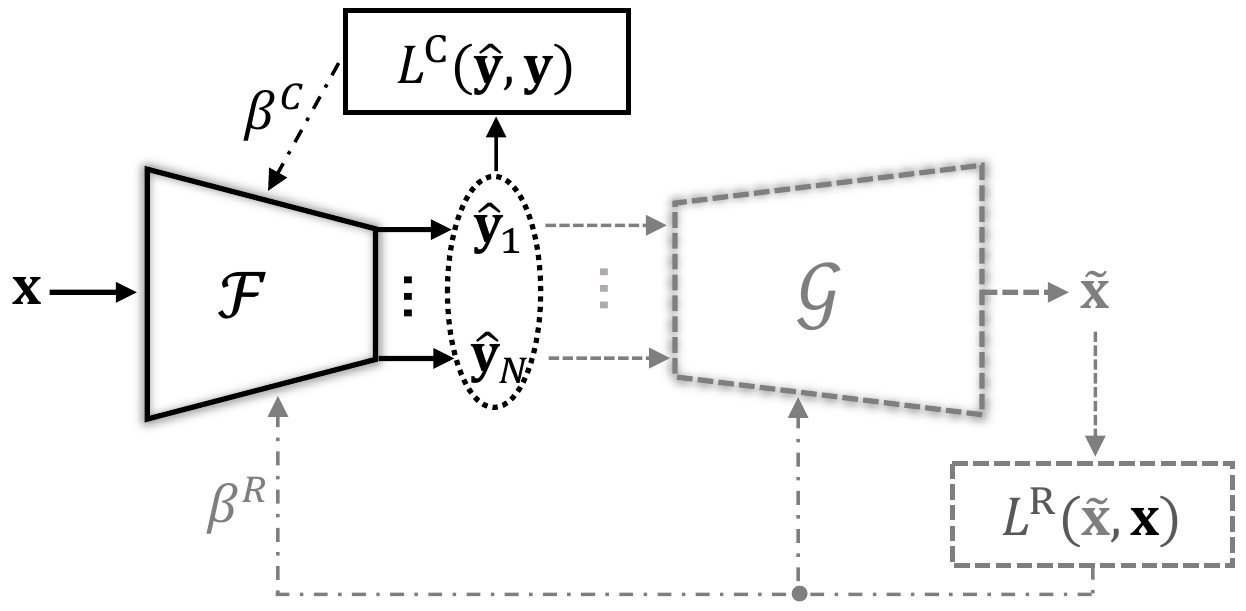}
    \caption{\small $\cF$ is the \pub model, $\cG$ is the attack model, $L^C$ is the classification loss, $L^R$ is the attack reconstruction loss. Both $\cF$ and $\cG$ are DNNs. Hyperparameters $\beta^{C}$ and $\beta^{R}$ control the trade-offs between classification and reconstruction tasks.}  
    \label{fig:model}
\end{wrapfigure}

{\bf Training of Target Classifier and Attack Models.} We present an algorithm for jointly training $\cF$ and $\cG$ (Figure~\ref{fig:model}). The \server trains model $\cF$ that takes data $\bx$ as input and produces $N$-outputs. Outputs are attached to the {\em classification loss function} $L^{C}$, which computes the amount of inaccuracy in predicting the true attribute $\by$, and thus provides gradients for updating $\cF$. For categorical attributes, we use the standard categorical cross-entropy loss 
\begin{equation}\label{eq_cce}
L^{C}(\byh, \by) =  \sum_{n=1}^{N} -\by_n \log \byh_n \text{,}
\end{equation}
and for binary attributes, we use the class-weighted binary cross-entropy
\begin{equation}\label{eq_bce}
L^{C}(\byh, \by) = \frac{1}{N}\sum_{n=1}^{N} \eta_n\by_n \log \byh_n + (1-\by_n) \log (1-\byh_n),
\end{equation}
where $\eta_n$ denotes the weight of class 1 for attribute $\by_n \in \{0,1\}$, and it is defined as the number of samples labeled $0$ divided by the number of samples labeled $1$ in the training dataset. CelebA dataset~\cite{liu2015celeba} used in our experiments is highly unbalanced for several attributes. Our motivation for using the class-weighted binary loss function is to obtain a fairer classification for unbalanced labels. While training $\cF$, the model's outputs are fed into another model $\cG$, which aims to reconstruct the original data. The output of $\cG$ is attached to a reconstruction loss function $L^{R}$, producing gradients for updating both $\cF$ and $\cG$.

In this paper, we benchmark image datasets in our experiments; thus, we utilize the loss functions used in image processing tasks~\cite{zhao2016loss}. In particular, we employ a weighted sum of $(i)$ {\em structural similarity index measure}~(SSIM)~\cite{wang2004image} and $(ii)$ {\em Huber loss}~\cite{hastie2009elements} which is a piecewise function including both mean squared error~(known as MSELoss) and mean absolute error~(MAE, also known as L1Loss)~\cite{paszke2019pytorch}. This design choice of combining a perceptually-motivated loss (\ie SSIM) with a statistically-motivated loss (\ie MSELoss or L1Loss) is inspired by the common practice used by previous work in image-processing literature~\cite{zhao2016loss, yoo2021sinir}:
\begin{equation}\label{eq:recon_loss_ssim_mse}
    L^{R}(\recx, \bx) = \alpha SSIM(\recx, \bx) + \gamma Huber(\recx, \bx),
\end{equation}
where $\alpha$ and $\gamma$ are the hyperparameters for data reconstruction. Note that, depending on the data type and the attack's purpose, one can use other reconstruction loss functions. 

{\bf The ultimate loss function.} For optimizing the parameters of $\cF$, we use:
\begin{equation}\label{eq:final_c_g_loss}
    L^{\cF} = \beta^{C}L^{C}(\byh, \by)+\beta^{R} L^{R}(\recx, \bx),
 \end{equation}
where $\beta^{C} $ and $\beta^{R}$ are the weights that allow us to move along different possible local minimas and both are non-negative real-valued. For optimizing the parameters of $\cG$, we only use $L^{R}$, but notice that there is an implicit connection between $\cG$, $\cF$, and $L^{C}$ since $\cG$ acts on $\byh=\cF(\bx)$. Algorithm~\ref{alg:training_f_g}~(Appendix~\ref{sec_apx_algo}) summarizes the explained training procedure.

\subsection{Reconstruction Risk}\label{sec_measure_rec_risk}
We define $\cS(\cdot,\cdot)$ as a measure of {\em reconstruction risk}, and $\cS(\recx,\bx) \geq \rr$ means  the risk of reconstructing $\bx$ is more than $\rr$ based on the measure $\cS$. Considering the {\em reconstruction} of data, $\recx$, we use $\cS$ to measure {\em privacy loss}. A pivotal question is: what is the most suitable and general $\cS$ for computing and evaluating the attacker's success? Previous works (see Appendix~\S\ref{sec_realted_work})  mostly rely on common measures such as MSE or SSIM. We propose our measure of reconstruction risk and we compare it with other measures in \S\ref{sec_exp_result}.

{\bf Basics.} For two random vectors $\bx$ and $\recx$ of the same distribution with covariance matrix $\covM$, the Mahalanobis distance~(MD) is a dissimilarity measure between $\bx$ and $\recx$:
%
$
    \MD(\bx,\recx) = \sqrt{(\bx-\recx)\covM^{-1}(\bx-\recx)}.
$
%
Similarly, we can compute $\MD(\bx,\meanD)$, where $\meanD$ is the mean of distribution that $\bx$ is drawn from. Notice that if $\covM$ is the identity matrix, MD reduces to the Euclidean distance (and thus the typical MSE). Notice that in practice, \eg for real-world data types such as images, the $\covM$ is rarely close to the identity matrix. The pixels of an image are correlated to each other. Similarly, the sample points of time-series signals are temporally correlated to each other, and so on. Therefore, for computing MD, we need to approximate $\covM$ using a sample dataset. In our experiments, we approximate $\covM$ via samples in the training dataset.  Another characteristic of MD is that when the data follows a  multivariate normal distribution, the probability density of an observation $\bx$ is uniquely determined by MD:
\begin{equation}
    \Pr(\bx) = \frac{1}{\sqrt{det(2\pi\covM)}}\exp\Big(-\frac{(\MD(\bx,\meanD)\big)^2}{2}\Big).
\end{equation}
The multivariate normal distribution is the most common probability distribution that is used for approximating data distribution~\cite{murphy2021probabilistic}. Thus, with the assumption that one can approximate data distribution using a multivariate normal distribution, MD can be utilized for computing the probability density of an observation $\bx$; \ie $\Pr(\bx)$.

{\bf Our Measure.} Motivated by the characteristics of MD, we assume that the risk of a reconstruction $\recx$ of a sample $\bx$ depends on both $\MD(\bx,\recx)$ and $\Pr(\bx)$: the more unlikely is a sample (lower $\Pr(\bx)$), the more important is the value of $\MD(\bx,\recx)$ (the more informative is a specific reconstruction). For the reconstructions of two independent samples $\bx^1$ and $\bx^2$ with $\MD(\bx^1,\recx^1)=\MD(\bx^2,\recx^2)$ and $\Pr(\bx^1)<\Pr(\bx^2)$, the risk of $\recx^1$ should be higher than $\recx^2$. Our intuition is: because $\bx^1$ is less likely than $\bx^2$, then $\bx^1$ is easier to be identified when attackers observe $\recx^1$, compared to $\bx^2$ when attackers observe $\recx^2$. Because $\bx^1$ is less likely (or more unique) than $\bx^2$, then a reconstruction of $\bx^1$ will give the attacker more information. 

In general, the intuition is that reconstructing a data point that belongs to a more sparse part of the population is riskier than reconstructing those that belong to a more dense part of the population. To this end, we define the {\em reconstruction risk} of a model with respect to a benchmark test dataset $\data=\{x^n\}_{n=1}^{N}$ as
\begin{equation}\label{eq_our_rec_risk}
    \rr = \frac{1}{N}\sum_{n=1}^{N}\MD(\bx^n,\meanD)\big/\MD(\bx^n,\recx^n).
\end{equation}
The less likely a sample, or the better its reconstruction quality, the higher is its contribution to the risk of the dataset. Our measure gives a general notion of reconstruction risk that depends on the characteristics of the entire dataset, and not just each sample independently. Moreover, our measure can be used across different data types and is not restricted to images or videos.   

{\em Remark.} Our proposed $\rr$ is task-agnostic. For example, for face images, background reconstruction might not be as important as eyes or mouth reconstruction. For task-specific risk assessments, one might need to perform preprocessing. For example, by image segmentation and applying the computation of $\rr$ only to that segment of the photo that includes the face.

\section{Experimental Results}\label{sec_exp_result}
\begin{table}[]
\centering
\caption{Reconstruction quality vs. classification accuracy in different settings and for different datasets. We repeat each experiment for five different random seeds and report the mean and standard deviation.}
\label{tab_results_main}
\resizebox{\textwidth}{!}{%
\begin{tabular}{@{}llc|cc|c|c@{}}
\toprule
Outputs & Dataset & $\beta^{R}/\beta^{C}$ & PSNR (dB) & SSIM & $\rr$ & ACC (\%) \\\midrule
\multirow{30}{*}{\textbf{Logits}}
& 
\multirow{5}{*}{MNIST} 
 & $0/1$ & 0.000   $\pm$   0.000 & 0.000  $\pm$  0.000 & 1.000  $\pm$  0.000 & 99.54  $\pm$  0.11 \\
 &   & $1/1$ & 22.090  $\pm$  0.095 & 0.920  $\pm$  0.003 & 1.378  $\pm$  0.021 & 99.53  $\pm$  0.09  \\
 &   & $3/1$ & 22.367  $\pm$  0.075 & 0.926  $\pm$  0.001 & 1.394  $\pm$  0.012 & 99.52  $\pm$  0.04 \\
 &   & $5/1$ & 22.336  $\pm$  0.049 & 0.927  $\pm$  0.001 & 1.392  $\pm$  0.011 & 99.55  $\pm$  0.04
 \\
 &   & $1/0$ & 22.169  $\pm$  0.052 & 0.926  $\pm$  0.001 & 1.398  $\pm$  0.009 & 10.00  $\pm$  0.00
 \\\cline{2-7}
 & \multirow{5}{*}{FMNIST} 
    & $0/1$ & 0.000  $\pm$  0.000 & 0.000  $\pm$  0.000 & 1.000  $\pm$  0.000 & 94.33  $\pm$  0.12 \\
 &  & $1/1$ & 20.525  $\pm$  0.056 & 0.783  $\pm$  0.001 & 1.243  $\pm$  0.003 & 94.58  $\pm$  0.05  \\
 &  & $3/1$ & 20.883  $\pm$  0.061 & 0.799  $\pm$  0.001 & 1.272  $\pm$  0.006 & 94.24  $\pm$  0.08 \\
 &  & $5/1$ & 20.871  $\pm$  0.068 & 0.803  $\pm$  0.001 & 1.271  $\pm$  0.004 & 94.30  $\pm$  0.16 \\
 &  & $1/0$ & 21.021  $\pm$  0.032 & 0.810  $\pm$  0.001 & 1.281  $\pm$  0.008 & 10.00  $\pm$  0.00
 \\\cline{2-7}
 & \multirow{5}{*}{CIFAR10}
 & $0/1$ & 0.000  $\pm$  0.000 & 0.000  $\pm$  0.000 & 1.000  $\pm$  0.000 & 91.86  $\pm$  0.69 \\
 &  & $1/1$ & 15.388  $\pm$  0.039 & 0.377  $\pm$  0.002 & 1.009  $\pm$  0.003 & 91.34  $\pm$  0.58 \\
 &  & $3/1$ & 15.550  $\pm$  0.042 & 0.406  $\pm$  0.001 & 1.013  $\pm$  0.001 & 90.84 $\pm$ 0.35 \\
 &  & $5/1$ & 15.581 $\pm$ 0.043 & 0.414 $\pm$ 0.003 & 1.010 $\pm$ 0.003 & 90.62 $\pm$ 0.62 \\
 &  & $1/0$ & 15.784 $\pm$ 0.037 & 0.468 $\pm$ 0.000 & 1.026 $\pm$ 0.002 & 10.00 $\pm$ 0.00
 \\\cline{2-7}
 & \multirow{5}{*}{CIFAR100} 
 & $0/1$ & 0.000 $\pm$ 0.000 & 0.000 $\pm$ 0.000 & 1.000 $\pm$ 0.000 & 68.13 $\pm$ 0.66 \\
 &  & $1/1$ & 16.757 $\pm$ 0.142 & 0.473 $\pm$ 0.012 & 1.051 $\pm$ 0.003 & 67.50 $\pm$ 0.66 \\
 &  & $3/1$ & 18.463 $\pm$ 0.311 & 0.646 $\pm$ 0.018 & 1.147 $\pm$ 0.023 & 64.57 $\pm$ 1.20 \\
 &  & $5/1$ & 19.047 $\pm$ 0.235 & 0.701 $\pm$ 0.008 & 1.201 $\pm$ 0.015 & 61.25 $\pm$ 1.34 \\
 &  & $1/0$ & 20.693 $\pm$ 0.097 & 0.821 $\pm$ 0.002 & 1.454 $\pm$ 0.011 & 1.00 $\pm$ 0.00 \\\cline{2-7}
 & \multirow{7}{*}{TinyImgNet} 
    & $0/1$ & 0.000 $\pm$ 0.000 & 0.000 $\pm$ 0.000 & 1.000 $\pm$ 0.000 & 46.96 $\pm$ 0.25 \\
 &  & $1/1$ & 16.763 $\pm$ 0.171 & 0.473 $\pm$ 0.014 & 1.042 $\pm$ 0.004 & 45.98 $\pm$ 0.56 \\
 &  & $3/1$ & 19.036 $\pm$ 0.190 & 0.692 $\pm$ 0.011 & 1.166 $\pm$ 0.016 & 42.71 $\pm$ 0.22 \\
 &  & $5/1$ & 20.072 $\pm$ 0.181 & 0.766 $\pm$ 0.009 & 1.261 $\pm$ 0.019 & 37.57 $\pm$ 1.39 \\
 &  & $1/0$ & 23.166 $\pm$ 0.104 & 0.900 $\pm$ 0.004 & 1.796 $\pm$ 0.028 & 0.50 $\pm$ 0.00
 \\\cline{2-7}
  & \multirow{5}{*}{CelebA} 
    & $0/1$ & 0.000 $\pm$ 0.000 & 0.000 $\pm$ 0.000 & 1.000 $\pm$ 0.000 & 88.43 $\pm$ 0.00 \\
 &  & $1/1$ & 19.081 $\pm$ 0.044 & 0.813 $\pm$ 0.000 & 1.238 $\pm$ 0.002 & 88.03 $\pm$ 0.06 \\
 &  & $3/1$ & 19.619 $\pm$ 0.087 & 0.837 $\pm$ 0.002 & 1.287 $\pm$ 0.009 & 86.41 $\pm$ 0.37 \\
 &  & $5/1$ & 19.850 $\pm$ 0.018 & 0.845 $\pm$ 0.000 & 1.308 $\pm$ 0.002 & 85.82 $\pm$ 0.20  \\
 &  & $1/0$ & 20.292 $\pm$ 0.021 & 0.858 $\pm$ 0.001 & 1.346 $\pm$ 0.002 & 50.00 $\pm$ 0.00 \\
\hline\hline
\multirow{10}{*}{\textbf{Softmax}}
& 
\multirow{2}{*}{MNIST} 
    & $1/1$ & 16.061 $\pm$ 0.186 & 0.684 $\pm$ 0.012 & 1.037 $\pm$ 0.006 & 99.54 $\pm$ 0.05 \\
 &  & $5/1$ & 21.036 $\pm$ 0.127 & 0.897 $\pm$ 0.004 & 1.272 $\pm$ 0.018 & 99.62 $\pm$ 0.04
 \\\cline{2-7}
 & \multirow{2}{*}{FMNIST} 
    & $1/1$ & 18.009 $\pm$ 0.243 & 0.684 $\pm$ 0.011 & 1.086 $\pm$ 0.010 & 94.38 $\pm$ 0.06 \\
 &  & $5/1$ & 19.915 $\pm$ 0.182 & 0.771 $\pm$ 0.006 & 1.204 $\pm$ 0.007 & 94.29 $\pm$ 0.35
 \\\cline{2-7}
 & \multirow{2}{*}{CIFAR10} 
    & $1/1$ & 13.934 $\pm$ 0.246 & 0.256 $\pm$ 0.031 & 1.003 $\pm$ 0.001 & 91.75 $\pm$ 0.38  \\
 &  & $5/1$ & 15.389 $\pm$ 0.061 & 0.409 $\pm$ 0.004 & 1.015 $\pm$ 0.001 & 90.47 $\pm$ 0.56
 \\\cline{2-7}
 & \multirow{2}{*}{CIFAR100} 
    & $1/1$ & 12.846 $\pm$ 0.125 & 0.202 $\pm$ 0.002 & 1.002 $\pm$ 0.001 & 67.21 $\pm$ 0.18 \\
 &  & $5/1$ & 16.580 $\pm$ 0.138 & 0.510 $\pm$ 0.002 & 1.054 $\pm$ 0.001 & 65.65 $\pm$ 1.21
 \\\cline{2-7}
 & \multirow{2}{*}{TinyImgNet} 
    & $1/1$ & 13.466 $\pm$ 0.042 & 0.203 $\pm$ 0.003 & 1.003 $\pm$ 0.000 & 44.08 $\pm$ 1.70 \\
 &  & $5/1$ & 17.168 $\pm$ 0.084 & 0.569 $\pm$ 0.010 & 1.066 $\pm$ 0.003 & 42.38 $\pm$ 0.68
 \\
\bottomrule
\end{tabular}%
}
\end{table}
Our experimental setup is detailed in Appendix~\S\ref{sec_exp_setup}. Our main results are reported in Tables~\ref{tab_results_main}~(and Appendix~Table~\ref{tab_results_diff_K}). We compare the {\em accuracy} of the \pub task and the {\em reconstruction quality} for different trade-offs. We consider two settings: during training the attack model, $\cG$ receives (i) the {\em logits} outputs of $\cF$, or (ii) the {\em softmax} outputs. To compare the trade-offs between {\em accuracy} and {\em reconstruction quality}, we also show two extremes in training $\cF$: {\em classification only} (when $\beta^R=0$) and  {\em reconstruction only} (when $\beta^C=0$).  The main findings are summarized as follows. 

Each {\em experiment} includes training $\cF$ and $\cG$ on the training dataset, for 50 epochs, and choosing $\cF$ and $\cG$ of the epoch in which we achieve the best result on the validation set according to loss function in Equation~\eqref{eq:final_c_g_loss}. In each experiment, via the test dataset, we evaluate $\cF$ by measuring the accuracy of $\cF$ in estimating the public task using Equations~\eqref{eq_pub_acc_cat}~or~\eqref{eq_pub_acc_bin}, and we evaluate $\cG$ by computing the reconstruction quality measured by PSNR, SSIM, and our proposed reconstruction risk $\rr$ in ~\eqref{eq_our_rec_risk}.  To compute  $\rr$, we approximate $\meanD$ and $\covM$ via samples in the training set. For a fair comparison, we use a random seed that is fixed throughout all the experiments, thus the same model initialization and data sampling are used. 

{\bf (1) Trade-offs.} When the logit outputs are available, the attacker can keep the classification accuracy very close to that achieved by a \hon model, while achieving a reconstruction quality close to that of reconstruction only.  For instance, even for relatively complex samples from TinyImgNet, we observe that with about $4\%$ loss in accuracy (compared to classification-only setting), we get a reconstruction quality of around $19$ dB PSNR and $0.7$ SSIM; which is not as perfect as reconstruction only but can be considered as a serious privacy risk. We observe serious privacy risks for other, less complex data types.  We do not perform any hyper-parameter or network architecture search (as it is not the main focus of our work). However, our results show that if one can perform such a search and find a configuration that achieves better performance (compared to our default WideResNet) in classification- and reconstruction-only settings, then such a model is also capable of achieving a better trade-off. Thus, our current results can be seen as a lower bound on the capability of an attacker. 

{\bf (2) Data Type.} For grayscale images (MNIST and FMNIST) we demonstrate very successful attacks. For colored images (CIFAR10, CIFAR100, TinyImgNet), it is harder to achieve as good trade-offs as those achieved for grayscale images. However, as one would expect, when the number of classes goes up, \eg from 10 to 100 to 200, the quality of reconstruction also becomes much better, \eg from PSNR of about 15 to about 18 to about 20 dB, for CIFAR10, CIFAR100, and TinyImgNet respectively.

{\bf (3) Logits vs. Softmax.} As one may expect, transforming logits into softmax probabilities will make it harder to establish a good trade-off. However, we still observe reasonably good trade-offs for low-complexity data types. The difficulty is more visible when the data complexity goes up. For TinyImgNet in the softmax setting, we can get almost the same reconstruction quality of the logit setting (around 17dB PSNR and 0.57 SSIM); however, the classification accuracy in the softmax setting drops by about $3\%$ to compensate for this.  Notice that for the CelebA dataset, we cannot transform the outputs into softmax as the attributes are binary. Instead, we can use the sigmoid function, which is a one-to-one function, and allow the \server to easily transform the received sigmoid outputs into logit outputs. Hence, the \server can train $\cF$ and $\cG$ in the logit setting, and after training just attach a sigmoid activation function to the output layer. The fact that the shortcoming of the sigmoid setting can be easily resolved by such a simple trick will facilitate such attacks as releasing sigmoid values might look less suspicious. As a side note, {\em softmax} functions, unlike sigmoids, are not one-to-one; since $softmax(x)=softmax(x+a)$ for all real-valued $a$. Thus, such a trick cannot be applied to categorical attributes with more than two classes, where \users might release the {\em softmax} outputs instead of raw ones. In such settings, a \server can replace categorical attributes of size $C$ with $C$ binary attributes. We leave the investigation of such a replacement to a future study. 

{\bf (4) The value of $N$.} For the CelebA, we observe that the reconstruction quality improves with the number of binary attributes $N$; however, the improvement is not linear. With $\beta^R/\beta^C=3$ we have about $0.14$ points improvement in SSIM and about $2$ dB in PSNR when going from $N=1$ to $N=5$ attributes, but when moving from $N=5$ to $N=10$ we observe an improvement of only 0.05 in SSIM and about $1$ db in PSNR. A similar diminishing increase happens also when we move from $N=10$ to $N=40$ attributes. In sum, these results suggest that the proposed attack achieves meaningful performance on CelebA with just a few outputs ($10 \leq N \leq 20$), and such scenarios of collecting a few binary attributes can lie within several applications provided by real-word ML service providers. 

\begin{figure*}
\begin{minipage}[t]{\textwidth}
    \offinterlineskip
    \includegraphics[trim={0.5cm .5cm 0.5cm .5cm}, clip, width=\textwidth]{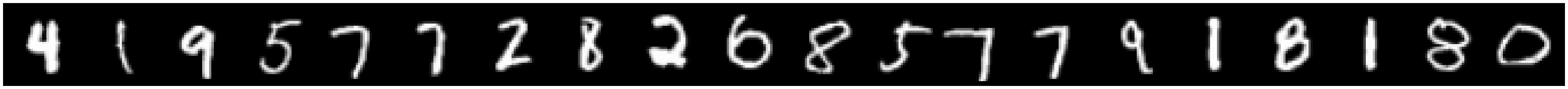}
    \includegraphics[trim={0.5cm .5cm 0.5cm .5cm}, clip, width=\textwidth]{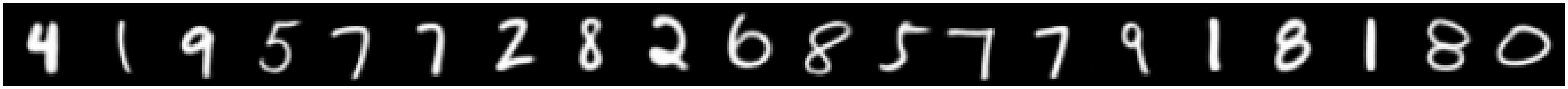}
    \offinterlineskip
    \includegraphics[trim={0.5cm .5cm 0.5cm .0cm}, clip, width=\textwidth]{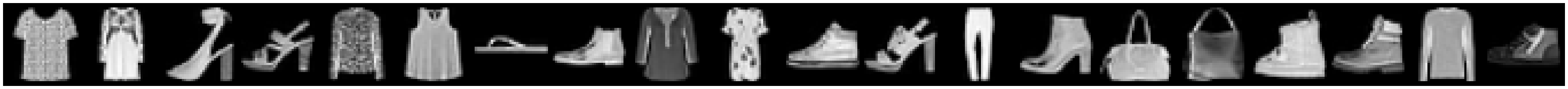}
    \includegraphics[trim={0.5cm .5cm 0.5cm .5cm}, clip, width=\textwidth]{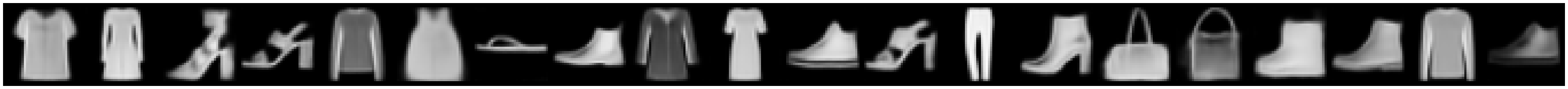}
    \offinterlineskip
    \includegraphics[trim={0.5cm .5cm 0.5cm .0cm}, clip, width=\textwidth]{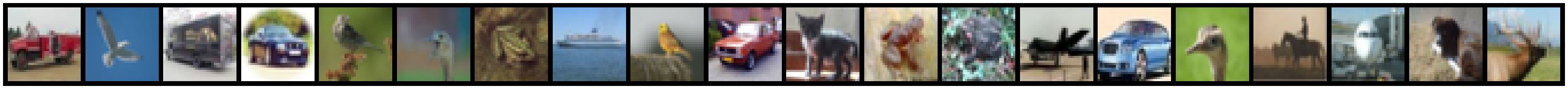}
    \includegraphics[trim={0.5cm .5cm 0.5cm .5cm}, clip, width=\textwidth]{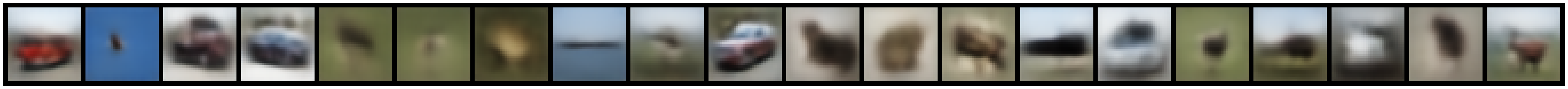}
    \offinterlineskip
    \includegraphics[trim={0.5cm .5cm 0.5cm .0cm}, clip, width=\textwidth]{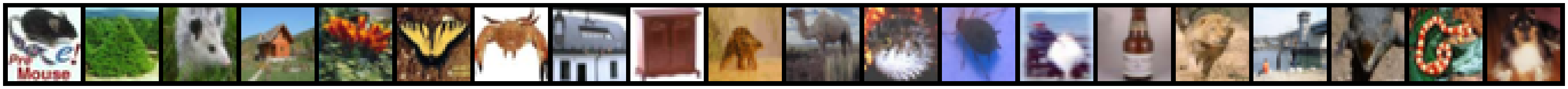}
    \includegraphics[trim={0.5cm .5cm 0.5cm .5cm}, clip, width=\textwidth]{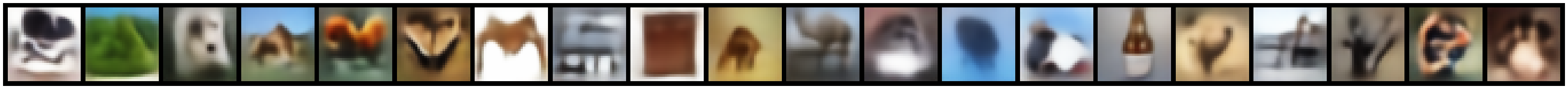}
    \offinterlineskip
    \includegraphics[trim={0.5cm .5cm 0.5cm .0cm}, clip, width=\textwidth]{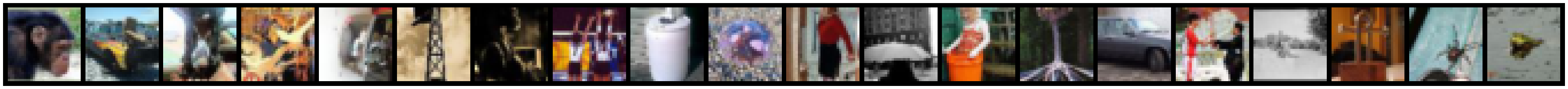}
    \includegraphics[trim={0.5cm .5cm 0.5cm .5cm}, clip, width=\textwidth]{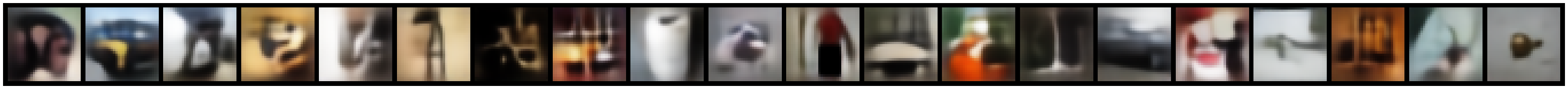}
    \offinterlineskip
    \includegraphics[trim={0.5cm .5cm 0.5cm .0cm}, clip, width=\textwidth]{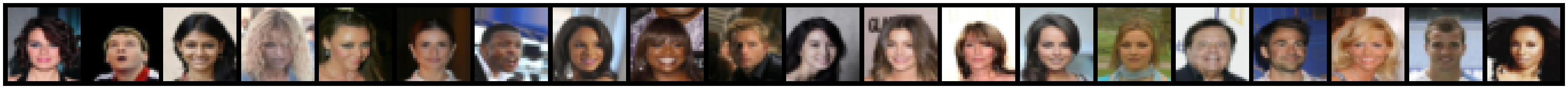}
    \includegraphics[trim={0.5cm .5cm 0.5cm .5cm}, clip, width=\textwidth]{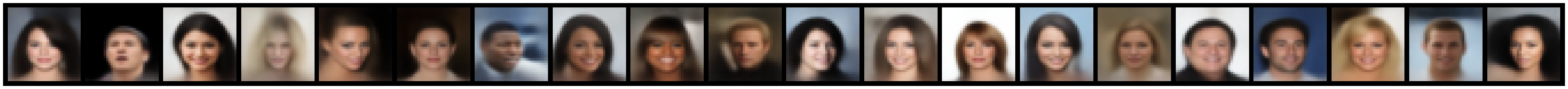}
\end{minipage}
\caption{Examples of image reconstruction with the logit outputs for $\beta^R/\beta^C= 3/1$ in Table~\ref{tab_results_main} for MNIST, FMNIST, CIFAR10, CIFAR100, TinyImageNet, and CelebA datasets (from top to bottom). For each dataset, the first row consists of the original images and the second row is the reconstructed data by the attacker.} \label{fig_qualitative_1}
\vspace{-0.3cm}
\end{figure*}

{\bf (5) The properties of $\rr$.}
We demonstrate the properties of
our proposed reconstruction risk $\rr$, compared to PSNR and SSIM; based on our main results reported in Table~\ref{tab_results_main}. The values of $\rr$ are in {\em conformity} with the values of PSNR and SSIM: the higher $\rr$ is, it indicates the higher PSNR and SSIM are. However, the values of PSNR and SSIM depend on the complexity of the data type, but the values of $\rr$ demonstrate more {\em homogeneity} across data types.  For example, in some situations, PSNR is almost the same but SSIM is different. For instance, with a similar PSNR of 19 dB for both CIFAR100 and CelebA, we observe different SSIM of 0.7 for CIFAR100 and 0.8 for CelebA). By combining both PSNR and SSIM, one can conclude that the model reveals more information on CelebA than CIFAR100. This can be seen by the values $\rr$ in which we have 1.23 for CelebA compared to 1.2 for CIFAR100. On the other hand, there are situations in which SSIM is almost the same but PSNR is different. For instance, SSIM of 0.81 for both FMNIST and CelebA corresponds to a PSNR of 21 dB for FMNIST and 19 dB for CelebA. Similarly, by combining these two measures, we expect the model to leak more information for FMNIST than CelebA. Again, this conclusion can be made by observing $\rr$ which is 1.28 for FMNIST and $1.23$ for CelebA.  Overall, $\rr$ provides a more consistent and unified measure of reconstruction risk, as it is based on a more general notion of distance than other data-specific measures~(see~\S\ref{sec_measure_rec_risk}).

{\bf (6) Qualitative comparison.}  Figure~\ref{fig_qualitative_1} (and similarly Appendix Figure~\ref{fig_qualitative_2}) show examples of data reconstruction. An interesting observation is that the reconstructed images are very similar to the original samples. We emphasize that in this paper we used off-the-self DNNs, and leave the design and optimization of dedicated DNN architectures to future studies.

\vspace{-0.5cm}
\section{Discussion}
\noindent {\bf Limitations.} Addressing all open questions when studying such a privacy risk is challenging. Throughout our evaluations, we decided to fix some variables, such as fixing data type to image data and the model architecture to WideResNet. We were motivated by the aim to allocate space and resources for thorough analysis and evaluation of more important variables, such as sample size, input and output complexity, hyper-parameters of the algorithms, potential defense, etc. Overall, our theoretical analysis, as well as the independence of our algorithm to input and model, suggests that similar results can be generalized to other data types and model architectures. 

\noindent{\bf Future Work.} (1) We only considered a single inference, but there are scenarios where multiple inferences are made on a user's private data; such as ensemble prediction using multiple models or Monte Carlo dropout. (2) Combining the outputs' of multiple models on the same data to improve the reconstruction quality is an open question. (3) We mainly focus on understanding the attack, and our initial effort on the defense is to inspire the community, to investigate more efficient and effective defenses. Our proposed defense mechanism needs several rounds of training and defining a threshold for attack detection. Considering defenses that potentially do not need training or can detect \vic models more accurately is also a key topic to explore. (4) Our focus is on classification; however, the foundational principles of our work apply to regression as well. Hence, one may consider exploring such potential attacks for regression models. (5) Finally, our proposed reconstruction risk can be further improved by comparing it with similar measures introduced for other attacks, such as the 'calibrated score' in~\cite{watson2022on} on membership inference attacks. 

\vspace{-0.5cm}
\section{Conclusion}
A growing paradigm in edge computing,  motivated by efficiency and privacy, is ``bringing the code to the data''. In this work, we challenge the privacy aspect of this paradigm by showing the possibility of data reconstruction from the outputs' of a \pub machine learning task. We benchmark data reconstruction risk and offer a unified measure for assessing the risk of data reconstruction. While detecting such a privacy attack is not trivial, we also take an initial step by proposing a practical technique for examining ML classifiers. We believe that our paper will serve as an inspiration for further explorations, in both attack and defense methods, to enhance data privacy in edge computing.

{\bf Acknowledgment.}
This work was partially supported by the UK EPSRC grant (grant no. EP/T023600/1 and EP/W035960/1) under the CHIST-ERA program, and received funding from the UKRI for the projects AI-R (ERC Consolidator Grant, EP/X030806/1) and INFORMED-AI (EP/Y028732/1). We would like to thank James Townsend for a fruitful discussion about this work.

\bibliography{refs}
\appendix
\section*{APPENDIX}
\section{Threat Model}\label{sec_threat_model}
{\bf Model's Utility.} 
Let $\bbI(\mathrm{C})$ denote the indicator function that outputs 1 if condition $\mathrm{C}$ holds, and 0 otherwise. Given a test dataset $\bbD^{test} \sim (\cX, \cY)$, we define the model's {\em accuracy} in performing the \pub task $\cY$ as either:
\begin{equation}\label{eq_pub_acc_cat}
    Acc \equiv \Pr_{(\bx, \by) \sim \bbD^{test},\text{ } \byh \leftarrow \cF(\bx)} \big[ \bbI\big(\argmax(\byh) = \by\big)\big],
\end{equation}
when $\by$ denotes a categorical attribute, or
\begin{equation}\label{eq_pub_acc_bin}
    Acc \equiv \frac{1}{N} \sum_{n=1}^{N} \Pr_{(\bx, \by_n) \sim \bbD^{test},\text{ } \byh_n \leftarrow \cF(\bx)} [ \bbI(\byh_n > \tau_n) = \by_n], 
\end{equation} 
when $\by_n$  denotes the $n$-th binary attribute and $\tau_n$ is the chosen threshold for that attribute. Usually, $\tau_n=0$ for the logit outputs and $\tau_n=0.5$ for softmax outputs (which is equivalent to sigmoid outputs for binary attributes). 

{\bf Data Privacy.}
We assume that a user shares the computed outputs with the \server, and this is the only information that the \server can get access to. The \server is supposed to perform only the \pub task and not infer any information other than the pre-specified \pub class/attributes. In particular, the \server should not be able to (approximately) reconstruct the \user's private data $\bx$.   We show the possibility that the \server can build an {\em attack function} $\cG$ for the {\em reconstruction} of the user's private data from the received outputs; \ie $\recx = \cG(\byh)$. We define $\cS(\cdot,\cdot)$ as a measure of {\em reconstruction risk}, and $\cS(\recx,\bx) \geq \rr$ means that the {\em risk of reconstructing $\bx$ is more than $\rr$ based on the measure $\cS$}.  Examples of such $\cS$ in the literature are PSNR for general signals and SSIM for images.  Thus, before measuring the privacy risk of a model, regarding data reconstruction attack, we first need to define $\cS$; as we do in our evaluations.  We say a classifier is {\em \vic} if it is trained with an objective function that captures private information (beyond the \pub task) in the model's output. We refer to models that are only trained for the \pub task as {\em \hon} models.

{\bf Curious vs. \Vic} As mentioned in \S\ref{sec_realted_work}, ~\cite{malekzadeh2021honest} shows that  ML models can be trained to be ``honest-but-curious'' and to reveal a \priv attribute just through the model's output for a \pub attribute. In our work, we coin the term {\em \vic} classifier to distinguish it from a {\em curious} classifier in~\cite{malekzadeh2021honest}. A {\em curious} classifier is trained to encode a single \priv attribute into the output of a model intended for another \pub attribute, and a {\em \vic} classifier aims to reconstruct the entire input data, and consequently might infer many \priv attributes. Moreover, for a {\em curious} classifier in~\cite{malekzadeh2021honest}, the untrusted server selects the \priv attributes that they want to infer a priori at training time, but for a {\em \vic} classifier there is no need to choose a specific \priv attribute at training time, as we reconstruct the whole input data. 

\section{Related Work}\label{sec_realted_work}
There are several attacks on data privacy in ML, such as property inference~\cite{melis2019exploiting}, membership inference~\cite{shokri2017membership,salem2018ml, carlini2019secret}, model inversion for data reconstruction~\cite{fredrikson2015model, an2022mirror}, or model poisoning~\cite{biggio2012poisoning, munoz2017towards, jagielski2018manipulating}. However, these attacks target ``training'' data, or they happen after multiple rounds of interactions, during which large amounts of information are shared with untrusted parties. Another line of work considers attacks on the gradients shared in collaborative learning~\cite{zhu2019deep}, which again focuses on the training time, and not the test time when no gradient is computed.

Regarding attacks at {\em test time}, \cite{song2019overlearning} showed that due to ``overlearning'', the internal representations extracted by DNN layers can reveal \priv attributes of the input data that might not even be correlated to the \pub attribute. But, in~\cite{song2019overlearning} the adversary is allowed to observe a subset of internal representations. Our work considers data privacy at test time, where the \priv input as well as all the internal representations are hidden, and the adversary has access only to the model's outputs.   

A recent work~\cite{malekzadeh2021honest} shows that when \users only release the model's output, ``overlearning'' is not a major concern as standard models do not reveal significant information about another \priv attribute. But  ML models can be trained to be ``honest-but-curious''~\cite{malekzadeh2021honest} and to secretly reveal a \priv attribute just through the model's output for a \pub attribute. While they show how to encode a single \priv attribute into the output of a model intended for another \pub attribute, in our work we show how by having access only to the model's outputs we can reconstruct the entire input data, and consequently infer many \priv attributes. An untrusted server in~\cite{malekzadeh2021honest} needs to decide the \priv attribute that they want to infer at training time, but in this paper, we do not need to choose a specific \priv attribute at training time, as we reconstruct the whole input data, and consequently allowing the attacker to extract numerous attributes. Thus, our work is more generalizable and complementary, rather than competitive to~\cite{malekzadeh2021honest}.

In~\cite{luo2020feature, jiang2022comprehensive}, vertical federated learning is considered,  where there are two parties, each observing a subset of the features for the same set of samples. An untrusted party can reconstruct the values of the \priv scalar features owned by the other party only from the output of a model that both parties jointly train. A generative model is used to estimate the values of the \priv features assuming that the malicious party can make multiple queries on the same data of the victim party. However, the problem setting and threat model of~\cite{luo2020feature, jiang2022comprehensive} are quite different from our work as instead of two collaborating agents, we consider servers and their \users, where each \user owns a single data point, and shares the output of the model only once. Moreover, in~\cite{luo2020feature, jiang2022comprehensive}, it is assumed that the trained model is under the adversary's control at test time, but in our work, we assume that training is already performed by the server on another dataset, and the model is under the \user's control at test time.

Overall, our work provides a thorough evaluation of how ML service providers can perform a strong attack on data privacy (\ie data reconstruction attack) at {\em inference time}, and such an attack is possible even in a highly restricted scenario, in which the server observes only the {\em outputs} that are aimed for the {\em \pub task}, and even when data is not just a scalar but it is {\em multi-dimensional}; such as face images. 

Recently, we identified a closely related work~\cite{yang2019neural} that performs joint training of a classifier and an attack model (similar to Algorithm~\ref{alg:training_f_g}). While we acknowledge that the attack discussed in this paper was first introduced in~\cite{yang2019neural}, we believe our work offers valuable new insights and improvements for understanding, assessing, and defending against the attack. Compared to~\cite{yang2019neural}, our work provides the following contributions. (1) We do not just rely on the L2 norm, instead, we explore a broader range of reconstruction losses, such as SSIM and Huber loss. (2) We analyze scenarios where either logits or softmax outputs could be shared with the server. (3) We conduct a quantitative analysis on more datasets to demonstrate how the complexity of data types and the number of classes affect the results. (4) We offer an information-theoretical analysis of the attack and provide deeper insights into the factors contributing to the attacker's success. (5) We introduce a new measure for reconstruction risk assessment based on the Mahalanobis distance, which offers a probabilistic view on data reconstruction attack as a more principled way to evaluate the success of an attack across models and datasets. (6) We propose a method to estimate if a model is vicious by checking if it deviates from the ideal solution for its target task, assuming it was only trained with the claimed objective. (7) Finally, we open-source our code to help the community conduct further analysis, benchmarking, and develop defenses against the described attack. 

\section{An Information-Theoretical View}\label{sec_inf_analysis}
The objective function in Equation~\eqref{eq:final_c_g_loss} can also be formulated as the optimization function
\begin{equation}
    \max \   \tI(\bx,\byh) \ \  \text{  \textit{subject to}} \ \argmax(\byh) = \by.
\end{equation}
The ``penalty method''~\cite{boyd2004convex} can be used to replace a constrained optimization problem with an unconstrained problem, whose solution ideally converges to the solution of the original constrained problem. 
The unconstrained formulation is
\begin{equation}\label{eq_unconst_inf}
    \max \   \tI(\bx,\byh)  + \beta \ \bbI(\argmax(\byh) = \by),
\end{equation}
where hyperparameter $\beta$ is called the ``penalty parameter''.

This formulation can be particularly appropriate when the \server does not know what type of reconstruction measure (SSIM, MSE, L1, etc.) will be more relevant for the attack. Hence, instead of training for a specific reconstruction measure, the \server can try to capture as much information about $\bx$ as it can into the model's output $\byh$ (the first term in Equation~\eqref{eq_unconst_inf}), and the server's only constraint is to keep the output informative for the \pub classification task (the second term in Equation~\eqref{eq_unconst_inf}). The indicator function in Equation~\eqref{eq_unconst_inf} is not easy to use by existing optimization tools, but one can re-write the formula via a surrogate function as
\begin{equation}
    \max \   \tI(\bx,\byh)  - \beta H_{\by}(\byh), 
\end{equation}
where instead of maximizing the indicator function we minimize the cross-entropy function $H_{\by}(\byh) = -\sum_{n=1}^{N}\by_n\log\byh_n$, as in Equation~\eqref{eq_cce}. This problem then is equivalent to
\begin{equation}
    \max \   \tH(\byh)-\tH( \byh | \bx) - \beta H_{\by}(\byh).
\end{equation}
As most of the classifiers in practice are deterministic functions, we have $\tH( \byh | \bx)=0$, and the optimization problem becomes
\begin{equation}\label{eq_entropy_cross_ent}
    \max \   \tH(\byh)  - \beta \ \tH_{\by}(\byh). 
\end{equation}
We know that  $\tH_{\by}(\byh) \geq \tH(\by | \byh)$, as one can prove it in the following way:
\begin{equation}
\begin{split}
    &\tH_{\by}(\byh) =   \tH(\by) - \tH(\by) + \tH_{\by}(\byh)
    \\
       &= -\sum_{n=1}^{N}\by_n\log\by_n + \sum_{n=1}^{N}\by_n\log\by_n - \sum_{n=1}^{N}\by_n\log\byh_n
    \\  &= -\sum_{n=1}^{N}\by_n\log\by_n + \sum_{n=1}^{N}\by_n\log(\frac{\by_n}{\byh_n})
    \\
    &= \tH(\by) + D_{KL}(\by || \byh) 
    = \tH(\by | \byh) + \tI(\by; \byh) + D_{KL}(\by || \byh) \geq \tH(\by | \byh),
    \end{split}
\end{equation}
where the last inequality is based on the fact that both mutual information and Kullback–Leibler divergence are non-negative.
Therefore, any solution for the optimization problem of \eqref{eq_entropy_cross_ent} provides a lower bound on the below formula
\begin{equation}
    \max \   \tH(\byh)  - \beta \ \tH(\by | \byh).  
\end{equation}

This sheds light on the attack's objective: {\em the attacker aims to maximize the entropy of the output while keeping the outputs informative about the \pub information}. Therefore, as long as the attacker keeps the outputs informative about the \pub task, they can use the rest of the information capacity (\ie the output's entropy) to capture other private information. Thus, a \vic model should produce higher-entropy outputs than an \hon model. Inspired by this fact, we introduce a defense mechanism to distinguish \vic models in \S\ref{sec_defense}.
\section{Algorithm}\label{sec_apx_algo}
Algorithm~\ref{alg:training_f_g} (in the appendix) summarizes the explained training procedure in \S\ref{sec_training}.

$(i)$ SSIM idea is that image pixels have strong inter-dependencies especially when they are spatially close. The SSIM index between image $\bx$ and its reconstruction $\recx$ is defined as: 
\begin{equation}\label{eq:ssim}
    SSIM(\recx, \bx) = \frac{\big(2\mu_{\bx}\mu_{\recx}+(k_1L)^2\big)\big(2\sigma_{\bx\recx}+(k_2L)^2\big)}{\big(\mu^2_{\bx}+\mu^2_{\recx}+(k_1L)^2\big)\big(\sigma_{\bx}+\sigma_{\recx}+(k_2L)^2\big)},
\end{equation}
where $\mu_{\bx}$ (similarly $\mu_{\recx}$), $\sigma_{\bx}$ (similarly $\sigma_{\recx}$), and $\sigma_{\bx\recx}$ denote the average, variance, and co-variance values, respectively. Here, $L$ denotes the size of the range of pixel values. We normalize the pixel values to the range $[0,1]$, and hence, $L=1$. We use the default values of $k_1=0.01$, $k_2=0.03$, and the default window size $11\times 11$. The SSIM in Equation~\eqref{eq:ssim}  is computed for every window, and the mean SSIM value among all the windows is considered as the final similarity index between two images. As SSIM lies within $[0,1]$, $SSIM(\recx,\bx) = 1$ is seen as the perfect reconstruction. 

$(ii)$ Huber loss computes the mean squared error~(MSE) if the absolute pixel-wise error falls below its parameter $\delta$; otherwise, it uses $\delta$-scaled mean absolute error (the default value for $\delta=1$). Thus, Huber loss combines advantages of both MSE and L1 losses; as the L1 term is less sensitive to outliers, and the MSE term provides smoothness near 0: 
\begin{equation}\label{eq:huber}
    Huber(\recx, \bx) = 
    \begin{cases} 
      0.5\text{MSELoss}(\recx , \bx) & \text{if } \text{L1Loss}({\recx , \bx}) < \delta \\
      \delta(\text{L1Loss}({\recx , \bx})-0.5\delta) & otherwise.
   \end{cases}
\end{equation}
The minimum value of Huber loss is 0 (\ie perfect reconstruction), and the maximum value depends on the input data dimensions. Thus, we need some hyperparameters to regularize the importance of structural similarity loss (SSIM) compared to the pixel-wise reconstruction loss (Huber). 

\begin{algorithm}[t]
\caption{Training classifier $\cF$ and attack model $\cG$.}\label{alg:training_f_g}
\begin{algorithmic}[1]
\STATE {\bfseries Input:} $\cF$: an $N-$output classifier, $\cG$: $N-$input data reconstruction model, $\bbD^{train}$: training set, $\bbD^{valid}$: validation set, $(\beta^{C}, \beta^{R})$: classification-reconstruction trade-off multipliers, $E$: number of epochs, $K$: batch size, $L^C$: classification loss, $L^R$: reconstruction loss, $OPT^{\cF}$ and $OPT^{\cG}$: optimizers for $\cF$ and $\cG$. 
\STATE {\bfseries Output:} trained classifier $\cF$ and attack model $\cG$, 
\STATE {\bf begin} 
\STATE $\cF,\cG = randomly\_initialize(\cF), randomly\_initialize(\cG)$
\STATE best\_validation\_result = 0 
\FOR{$e: 1,\ldots, E$} 
\FOR{$b: 1,\ldots, |\bbD^{train}|/K$} 
\STATE $Loss^{C} = \text{ }Loss^{R} = 0$
\FOR{$k: 1,\ldots, K$}
\STATE $(\bx, \by)\sim\bbD^{train}$ $\#$ random uniform sampling
\STATE $\byh = \cF(\bx)$  
\STATE $Loss^{C} = Loss^{C} + L^{C}(\byh, \by)$ $\#$ Equation~\eqref{eq_cce} or~\eqref{eq_bce}
\STATE $\recx = \cG(\byh)$ 
\STATE $Loss^{R} = Loss^{R} + L^{R}(\recx, \bx)$ $\#$ Equation~\eqref{eq:recon_loss_ssim_mse}
\ENDFOR
\STATE $Loss^{C}= Loss^{C} / (K)$
\STATE $Loss^{R} = Loss^{R} / (K)$ 
\STATE $\cF = OPT^{\cF}(\beta^{C} Loss^{C}+\beta^{R} Loss^{R})$ $\#$ Equation~\eqref{eq:final_c_g_loss}
\STATE $\cG = OPT^{\cG}(\beta^{R} Loss^{R})$ 
\ENDFOR
\STATE result = $evaluate(\cF, \cG, \beta^{C}, \beta^{R}, \bbD^{valid})$ $\#$
\IF{result $\geq$ best\_validation\_result}   
\STATE best\_validation\_result = result 
\STATE $\cF',\text{ } \cG' = \cF,\text{ } \cG $
\ENDIF
\ENDFOR
\RETURN  $\cF' \text{ and } \cG'$
\end{algorithmic}
\end{algorithm}

\section{Experimental Setup}\label{sec_exp_setup}
\subsection{Datasets}
Table~\ref{tab_datasets} lists the datasets we use for evaluations. We use benchmark datasets of different complexity to measure the performance of the attackers in different scenarios. MNIST, FMNIST, and CIFAR10 all have 10 classes, but the complexity of their data types are quite different. CIFAR10, CIFAR100, and TinyImgNet are of (almost) the same data type complexity, but their number of classes is different.
CelebA is a dataset including more than 200K celebrity face images, each with 40 binary attributes, \eg the `Smiling' attribute with values $0$:$Non\text{-}Smile$ or $1$:$Smile$. 
Just a few of these attributes are balanced (having almost equal numbers of 0s and 1s). For the results reported in Table~\ref{tab_results_diff_K}, when we want to choose $K$ attributes, we choose the $K$ most balanced ones. For a fair comparison among datasets, we reshape all the images to the same width and height of $32$. All datasets are split into training and test sets, by the publishers. For the validation set during training, we randomly choose 10\% of training data and perform the training only on the remaining 90\%; except for CelebA, in which the publisher has already included a validation set. 

\begin{table}[t]
\centering
\caption{Datasets used in our experiments.}
\label{tab_datasets}
\centering
\resizebox{.8\columnwidth}{!}{%
\begin{tabular}{@{}lllll@{}}
\toprule
Labels & Dataset~(ref.) & Classes & Shape & Samples \\ \midrule
\multirow{5}{*}{Categorical} & MNIST~\cite{lecun1998mnist} & 10 & 32$\times$32$\times$1 & 60K \\
 & FMNIST~\cite{xiao2017fashion} & 10 & 32$\times$32$\times$1 & 60K \\
 & CIFAR10~\cite{krizhevsky2009learning} & 10 & 32$\times$32$\times$3 & 50K \\
 & CIFAR100~\cite{krizhevsky2009learning} & 100 & 32$\times$32$\times$3 & 50K \\
 & TinyImgNet~\cite{le2015tiny} & 200 & 32$\times$32$\times$3 & 100K \\\hline
Binary & CelebA~\cite{liu2015celeba} & 40 & 32$\times$32$\times$3 & 200K \\ \bottomrule
\end{tabular}%
}
\end{table}

\subsection{Models and Training}\label{sec:our_dnn}
For $\cF$ and $\cG$ in Figure~\ref{fig_bigpic}, we borrow one of the commonly-used architectures for data classification: WideResNet~\cite{Zagoruyko2016WRN}. WideResNet allows decreasing the depth of the model (i.e., the number of layers) and instead increases the width of the residual networks; which has shown better performance over other thin and very deep counterparts. For $\cF$, we use WideResNet of width 5 which has around 9M parameters. For $\cG$, we use an architecture similar to that of $\cF$, but we replace the convolutional layers with transpose-convolutional layers.  For all the experiments, we use a batch size of 250 images, and Adam optimizer~\cite{kingma2014adam} with a learning rate of $0.001$. 

\subsection{Settings} Each {\em experiment} includes training $\cF$ and $\cG$ on the training dataset based on Algorithm~\ref{alg:training_f_g}, for 50 epochs, and choosing $\cF$ and $\cG$ of the epoch in which we achieve the best result on the validation set according to loss function in Equation~\eqref{eq:final_c_g_loss}. In each experiment, via the test dataset, we evaluate $\cF$ by measuring the accuracy of $\cF$ in estimating the public task using Equations~\eqref{eq_pub_acc_cat}~or~\eqref{eq_pub_acc_bin}, and we evaluate $\cG$ by computing the reconstruction quality measured by PSNR, SSIM, and our proposed reconstruction risk $\rr$ in ~\eqref{eq_our_rec_risk}.  To compute  $\rr$, we approximate $\meanD$ and $\covM$ via samples in the training set. For a fair comparison, we use a random seed that is fixed throughout all the experiments, thus the same model initialization and data sampling are used.

\section{Additional Results}

The fine-tuning of hyperparameters is a task at training time, and a \server with enough data and computational power can find near-optimal values for hyperparameters, as we do here using the validation set. 
In Appendix (Tables~\ref{tab_add_results_1} and  \ref{tab_add_results_3}) we report the results of additional experiments to show the effect of other hyperparameters on the established trade-offs.

In Tables~\ref{tab_add_results_1} and  \ref{tab_add_results_3}, we report the results of additional experiments to show the effect of some hyperparameter tuning on the established trade-offs. Table~\ref{tab_add_results_1} shows that using both Huber and SSIM allows the attacker to achieve better accuracy while keeping the reconstruction quality similar to other choices of loss function. In Table~\ref{tab_add_results_3}, we report the results when choosing different $\delta$ values for the Huber loss function. The lower $\delta$ is, we achieve the better accuracy but with a cost of lower reconstruction quality. 

\begin{table}[h]
\caption{Choosing different objective functions as the reconstruction Loss $L^R$ in \eqref{eq:final_c_g_loss}. The experiment is on TinyImgNet dataset when releasing the logit outputs.}
\label{tab_add_results_1}
\centering
\resizebox{.8\columnwidth}{!}{%
\begin{tabular}{@{}lccccc@{}}
\toprule
Loss & $\beta^R/\beta^C$ & PSNR & SSIM &  $\rr$ &  ACC \\ \midrule
\multirow{2}{*}{Only MSE}  &   3/1 & 19.39 $\pm$ 0.07 & 0.67 $\pm$ 0.0023 & 1.15 $\pm$ 0.0024 & 34.37 $\pm$ 0.43 \\
& 1/0 & 24.34 $\pm$ 0.05 & 0.80 $\pm$ 0.0030 & 1.87 $\pm$ 0.0238 & 0.50 $\pm$ 0.0 \\\cline{2-6}
 \multirow{2}{*}{Only Huber}  & 3/1 & 18.71 $\pm$ 0.06 & 0.65 $\pm$ 0.0070 & 1.12 $\pm$ 0.0065 & 35.93 $\pm$ 0.32 \\
 & 1/0 & 24.05 $\pm$ 0.30 & 0.79 $\pm$ 0.0133 & 1.78 $\pm$ 0.0950 & 0.50 $\pm$ 0.0  \\\cline{2-6}
\multirow{2}{*}{Huber \& SSIM} &  3/1 & 19.02 $\pm$ 0.18 & 0.69 $\pm$ 0.0104 & 1.16 $\pm$ 0.0142 & 42.43 $\pm$ 0.64 \\
& 1/0 & 23.17 $\pm$ 0.10 & 0.90 $\pm$ 0.004 & 1.80 $\pm$ 0.028 & 0.50 $\pm$ 0.00 \\
\bottomrule
\end{tabular}%
}
\end{table}


\begin{table}[h]
\caption{Choosing different $\delta$ values for Huber loss in \eqref{huber}. The experiment is on the TinyImgNet dataset when releasing the logit outputs and $\beta^R/\beta^C=3$.}  
\label{tab_add_results_3}
\centering
\resizebox{.8\columnwidth}{!}{%
\begin{tabular}{@{}lcccc@{}}
\toprule
$\delta$ & PSNR & SSIM & $\rr$ & ACC \\ \midrule
0.01 &  17.553 $\pm$ 0.206 & 0.652 $\pm$ 0.013 & 1.110 $\pm$ 0.015 & 44.19 $\pm$ 0.50 \\
0.05 & 17.844 $\pm$ 0.084 & 0.647 $\pm$ 0.002 & 1.112 $\pm$ 0.006 & 44.06 $\pm$ 0.05 \\
0.10 & 18.227 $\pm$ 0.040 & 0.660 $\pm$ 0.006 & 1.128 $\pm$ 0.003 & 43.53 $\pm$ 0.32 \\
0.25 &  18.672 $\pm$ 0.117 & 0.678 $\pm$ 0.010 & 1.146 $\pm$ 0.011 & 41.28 $\pm$ 0.80 \\
0.50 &  18.974 $\pm$ 0.137 & 0.693 $\pm$ 0.001 & 1.164 $\pm$ 0.011 & 42.17 $\pm$ 0.55 \\
1.00 &  19.030 $\pm$ 0.036 & 0.689 $\pm$ 0.001 & 1.163 $\pm$ 0.000 & 42.74 $\pm$ 0.20
 \\
\bottomrule
\end{tabular}%
}
\end{table}

\begin{figure*}
\begin{minipage}[t]{\textwidth}
    \offinterlineskip
    \includegraphics[trim={0.5cm .5cm 0.5cm .5cm}, clip, width=\textwidth]{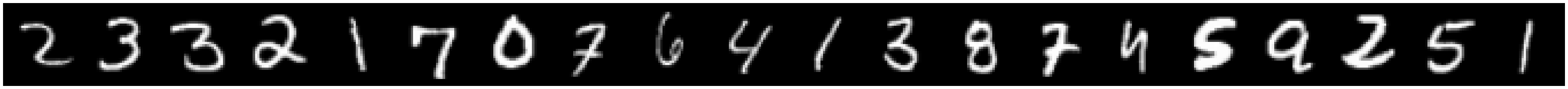}
    \includegraphics[trim={0.5cm .5cm 0.5cm .5cm}, clip, width=\textwidth]{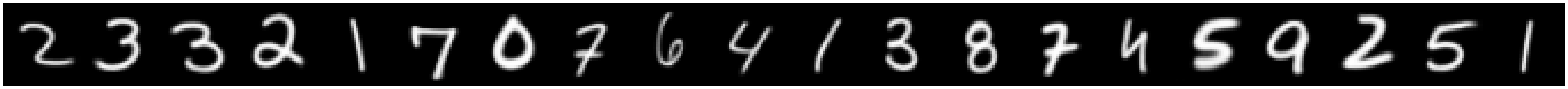}
    \offinterlineskip
    \includegraphics[trim={0.5cm .5cm 0.5cm .5cm}, clip, width=\textwidth]{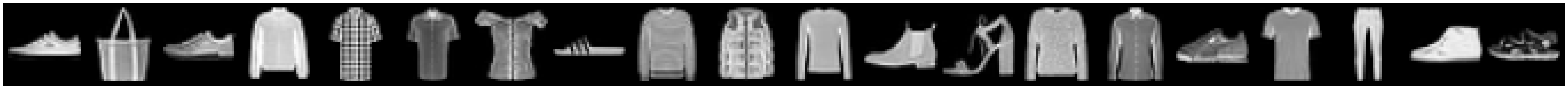}
    \includegraphics[trim={0.5cm .5cm 0.5cm .5cm}, clip, width=\textwidth]{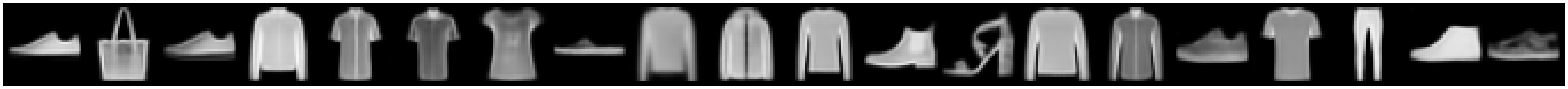}
    \offinterlineskip
    \includegraphics[trim={0.5cm .5cm 0.5cm .5cm}, clip, width=\textwidth]{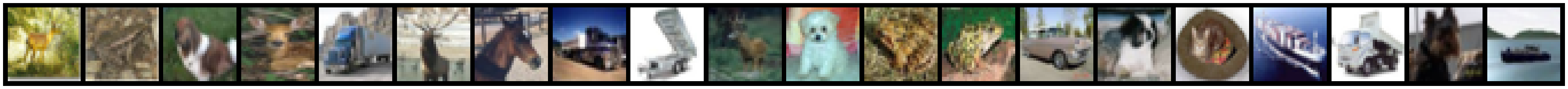}
    \includegraphics[trim={0.5cm .5cm 0.5cm .5cm}, clip, width=\textwidth]{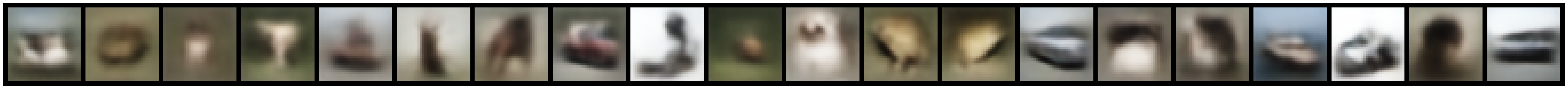}
    \offinterlineskip
    \includegraphics[trim={0.5cm .5cm 0.5cm .5cm}, clip, width=\textwidth]{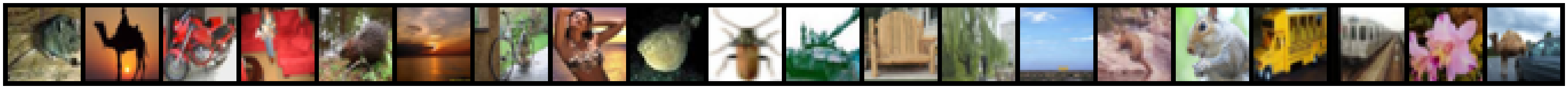}
    \includegraphics[trim={0.5cm .5cm 0.5cm .5cm}, clip, width=\textwidth]{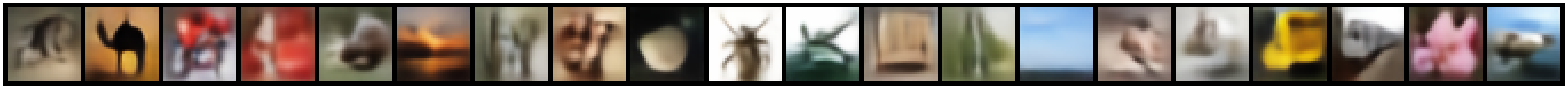}
    \offinterlineskip
    \includegraphics[trim={0.5cm .5cm 0.5cm .5cm}, clip, width=\textwidth]{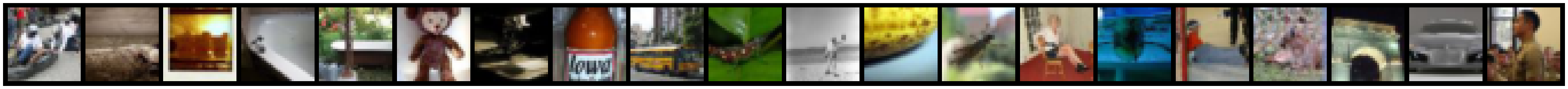}
    \includegraphics[trim={0.5cm .5cm 0.5cm .5cm}, clip, width=\textwidth]{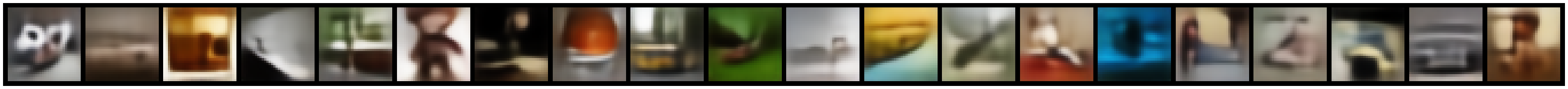}
    \offinterlineskip
    \includegraphics[trim={0.5cm .5cm 0.5cm .5cm}, clip, width=\textwidth]{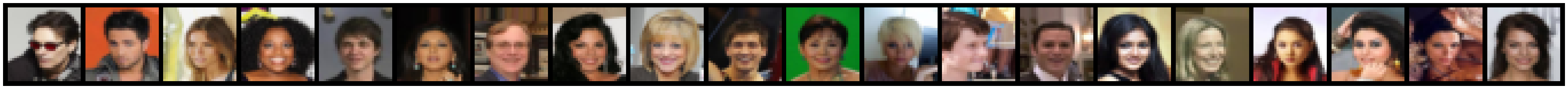}
    \includegraphics[trim={0.5cm .5cm 0.5cm .5cm}, clip, width=\textwidth]{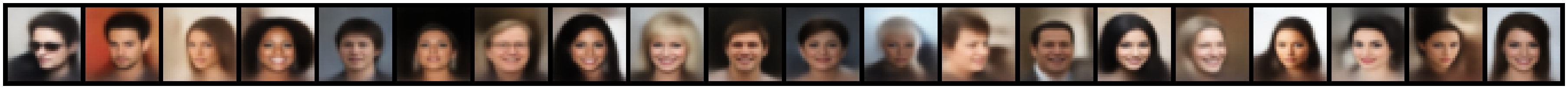}
\end{minipage}
\caption{More qualitative examples of the same experimental settings presented in Figure~\ref{fig_qualitative_1}.}\label{fig_qualitative_2}
\end{figure*}

\section{CelebA Dataset Details}
\begin{table}[h]
\caption{The code and name of face attributes in CelebA. We show the percentage (\%) of images having each attribute~(1s).}
\label{tab_celeba_code_name}
\centering
\resizebox{.8\columnwidth}{!}{%
\begin{tabular}{@{}llc|llc@{}}
Code & Attribute Name & \% of 1s & Code & Attribute Name & \% of 1s\\ \midrule
0 & 5oClockShadow & 12.4 & 20 & Male & 43.2 \\
1 & ArchedEyebrows & 27.5 & 21 & MouthSlightlyOpen & 50.3\\
2 & Attractive & 52.0 & 22 & Mustache & 03.4\\
3 & BagsUnderEyes & 21.7 & 23 & NarrowEyes & 12.5 \\
4 & Bald & 02.2 & 24 & NoBeard & 82.7 \\
5 & Bangs & 14.9 & 25 & OvalFace & 25.9 \\
6 & BigLips & 24.7 & 26 & PaleSkin & 04.3 \\
7 & BigNose & 24.8 & 27 & PointyNose & 28.1 \\
8 & BlackHair & 25.4 & 28 & RecedingHairline & 07.6 \\
9 & BlondHair & 14.8 & 29 & RosyCheeks & 07.0\\
10 & Blurry & 05.0 & 30 & Sideburns & 05.7\\
11 & BrownHair & 19.9 & 31 & Smiling & 51.0 \\
12 & BushyEyebrows & 14.8 & 32 & StraightHair & 21.8 \\
13 & Chubby & 06.5 & 33 & WavyHair & 28.7 \\
14 & DoubleChin & 04.5 & 34 & WearingEarrings & 20.7 \\
15 & Eyeglasses & 06.5 & 35 & WearingHat & 04.4 \\
16 & Goatee & 06.0 & 36 & WearingLipstick & 46.1 \\
17 & GrayHair & 05.4 & 37 & WearingNecklace & 12.8 \\
18 & HeavyMakeup & 38.2 & 38 & WearingNecktie & 07.9 \\
19 & HighCheekbones & 46.3 & 39 & Young & 75.4 \\ \bottomrule
\end{tabular}%
}
\end{table}
Table~\ref{tab_celeba_code_name} shows the code and name of 40 face attributes that are used in CelebA. Just a few of these attributes are balanced (having almost equal numbers of 0s and 1s) and some of them are highly correlated. For instance, if BlackHair is 1, then we certainly know that the other four attributes: Bald, BlondHair, BrownHair, and GrayHair are all 0.  Or, if Mustache is 1, then Male must be 1 as well. Some balanced attributes are Attractive, HeavyMakeup, Male, Smiling, and WavyHair. For the results reported in Table~\ref{tab_results_diff_K}, when we want to choose $K$ attributes, we choose the $K$ most balanced ones, as it is harder to train and evaluate classifiers for highly unbalanced attributes, and the focus of our paper is not on this aspect.

\begin{table}[]
\centering
\caption{Reconstruction risk vs. accuracy for CelebA dataset and different number of binary attributes.}
\label{tab_results_diff_K}
\resizebox{0.6\columnwidth}{!}{%
\begin{tabular}{@{}cccccc@{}}
\toprule
$N$ & $\beta^{R}/\beta^{C}$ & PSNR & SSIM & $\rr$ & ACC \\ \midrule
\multirow{2}{*}{1} & $1/1$ & 13.40 & 0.5112 & 1.0027 & 89.05 \\
   &  $3/1$ & 13.60 & 0.5231 &  1.0021 & 54.73 \\\cline{2-6}
\multirow{2}{*}{2} & $1/1$ & 13.68 & 0.5392 & 1.0089 & 88.55 \\
   & $3/1$ & 13.68 & 0.5292 & 1.0032 & 62.56 \\\cline{2-6}
\multirow{2}{*}{3} & $1/1$ & 14.17 & 0.5856 & 1.0248 &  91.98 \\
& $3/1$ & 14.49 & 0.6002 &  1.0343 & 86.65 \\\cline{2-6}
\multirow{2}{*}{4} & $1/1$ & 14.58 & 0.6015 & 1.0321 & 88.77  \\
& $3/1$ & 14.88 & 0.6232 & 1.0422 & 83.42 \\\cline{2-6}
\multirow{2}{*}{5} & $1/1$ & 14.72 & 0.6141 & 1.0370 &  87.85 \\
   & $3/1$ & 15.39 & 0.6543 & 1.0605 & 85.53 \\\cline{2-6}
\multirow{2}{*}{6} & $1/1$ & 15.14 & 0.6401 &  1.0530 & 87.60 \\
& $3/1$ & 15.71 & 0.6739 & 1.0745 & 86.27 \\\cline{2-6}
\multirow{2}{*}{8} & $1/1$ & 15.61 & 0.6710 & 1.0695 & 86.71 \\
& $3/1$ & 16.01 & 0.6938 & 1.0861 & 85.27 \\\cline{2-6}
\multirow{2}{*}{10} & $1/1$ & 15.83 & 0.6822 & 1.0768 & 86.25 \\
& $3/1$ & 16.33 & 0.7083 & 1.0992 & 85.55 \\\cline{2-6}
\multirow{2}{*}{15} & $1/1$ & 16.72 & 0.7239 & 1.1118 & 81.98 \\
& $3/1$ & 17.27 & 0.7507 & 1.1415 & 80.22 \\\cline{2-6}
\multirow{2}{*}{20} & $1/1$ & 17.35 & 0.7487 & 1.1415 & 82.03  \\
& $3/1$ & 17.89 & 0.7736 & 1.1713 & 80.62 \\\cline{2-6}
\multirow{2}{*}{30} & $1/1$ & 18.37 & 0.7874 & 1.1932 & 85.18  \\
& $3/1$ & 18.82 & 0.7994 & 1.2099 &  85.25 \\\cline{2-6}
\multirow{2}{*}{40} & $1/1$ & 19.13 & 0.8125 & 1.2358 & 88.09 \\
& $3/1$ & 19.70 & 0.8388 & 1.2954 & 86.78 \\
\bottomrule
\end{tabular}%
}
\end{table}

\section{Accuracy of Multi-Task Models}

Training the same model for multiple tasks concurrently, by assigning a dedicated output to each task is one of a kind in multi-task learning~(MTL)~\cite{caruana1997multitask}. There is empirical evidence that the accuracy of such an MTL model for each task can be very close to the setting where a dedicated model was trained for that task, and sometimes MTL could even achieve better accuracy due to learning representations that are shared across several tasks~(see these recent surveys on MTL~\cite{zhang2021survey, crawshaw2020multi}). 
In Appendix, Table~\ref{tab:resnet_vs_cnn}, we perform an experiment that shows training an MTL model on CelebA dataset can achieve very close accuracy to a counterpart models trained only for single tasks.

Training the same model for multiple tasks concurrently, by assigning a dedicated output to each task is one of a kind in multi-task learning~(MTL)~\cite{caruana1997multitask}. In Table~\ref{tab:resnet_vs_cnn}, for the 10-most-balanced attributes in CelebA, we compare the accuracy of a ResNet-50~\cite{he2016deep} (as a general-purpose model) with a simpler  convolutional neural network (CNN) proposed in~\cite{zhang2017age} for processing face images. The accuracy of CNN for face images (with about 7M parameters) is very close to what we can get from a ResNet-50 which has 3 times more parameters (25M parameters). Also, the CNN trained for all the $N=10$ tasks, in an MTL manner, achieves very close accuracy to when we train either model only for each task separately. Our motivation for showing these results is that having MTL models makes it easier for the \server to perform the reconstruction attack. While keeping the performance of the \pub task very close to standard models, the \server can exploit the capability of being MTL to learn more discerning and useful features, which further facilitate the malicious task; as shown by our main results.  

\begin{table}[h]
\caption{Test accuracy (\%) for the 10 most-balanced attributes in CelebA achieved by ResNet50 and the typical DNN used in this paper. Single-Task: the model has one output and it is trained only for that task. Multi-Task: the model has $N$ outputs jointly trained for all tasks.}
\label{tab:resnet_vs_cnn}
\centering
\resizebox{.8\columnwidth}{!}{%
\begin{tabular}{@{}lccc@{}}
\multicolumn{1}{l}{} & \multicolumn{3}{c}{Test Accuracy} \\ \cmidrule(l){2-4} 
\multicolumn{1}{l}{} & \multicolumn{1}{l}{ResNet50} & \multicolumn{2}{c}{CNN for Face Images} \\ \cmidrule(l){3-4} 
Attribute & Single-Task &Single-Task & Multi-Task ($N=10$) \\ \midrule
2:  Attractive & 80.05 & 80.11 & 79.93 \\
8:  BlackHair & 87.64 & 87.54 & 85.34 \\ 
11: BrownHair & 83.46 & 85.51 & 84.20 \\ 
18: HeavyMakeup & 89.63 & 89.53 & 89.54 \\ 
19: HighCheekbones & 86.17 & 86.45 & 86.02 \\ 
20: Male & 97.17 & 97.47 & 97.45 \\ 
31: Smiling & 91.82 & 92.22 & 92.00 \\ 
33: WavyHair & 77.14 & 80.41  & 79.45 \\ 
36: WearingLipstick & 92.25 & 92.92 & 93.38 \\ 
39: Young & 84.55 &  85.43  &  85.40 \\ 
\bottomrule
\end{tabular}%
}
\end{table}

\section{Potential Defense}\label{sec_defense}

There are some basic methods for defending against data reconstruction attacks. For example, defense mechanisms based on the {\em quantization} or {\em randomization} of the outputs before sharing them with the \server. Let us assume that the model's output for the $n$-th class/attribute is $\byh_n=0.87652345$. Such an output cannot be confidently explained, due to the complex nature of DNN models. However, one might hypothesize that the lower precision digits are probably exploited by the DNN to encode useful data for the \vic task of data reconstruction. Hence, one can propose to truncate/round the output and only release the $k$ higher precision digits~\cite{luo2020feature}; \eg to release $\byh_n=0.876$, $\byh_n=0.88$, or even $\byh_n=0.9$. 

On the other hand, one can randomize the output by adding zero-mean Laplacian/Gaussian noise with a predefined variance, similar to differential privacy mechanisms~\cite{dwork2014algorithmic}; \eg to release $\byh_n=0.87652345+\mathcal{N}(\mu=0,\sigma=.1)$. It is clear that these basic mechanisms provide some protection and will make the data reconstruction harder. The larger the noise variance or the number of truncated digits, the less successful must be the data reconstruction attack. However, these basic mechanisms also damage the utility of the \pub task, depending on the nature of the underlying service and the chosen parameters for quantization/randomization.

\subsection{Entropy Analysis}

The data $\bx$ is typically stored in a finite-precision floating point, but the entropy of $\bx$ could still be almost infinite: $\tH(\bx)\approx \infty$. Although the dimensionality of outputs is much lower than the dimensionality of inputs, it is the same for the entropy of the model's output: $\tH(\byh) \approx \infty$. Theoretically, without any constraint on the model and the output, $\byh$ could be shaped to carry almost all information needed to reconstruct $\bx$. Now, we explore the situations where there are some constraints on  $\byh$.

{\bf 1) Restricting the index of the max value}. The attacker should shape the output such that $\argmax(\byh)$ is accurate in predicting the \pub task $\by$. This constraint, by itself, does not significantly reduce the capability of the attacker. An attacker can build a model that allocates the largest possible floating point value to $\argmax(\byh)$ and thus freely use the rest of $N-1$ values. Considering $|\hat{\mathcal{Y}_l}|$ as the size of each logit output's alphabet, the entropy of this restricted output will be $\log( |\hat{\mathcal{Y}_l}|^{N-1})$, which is not much lower than the unrestricted case $\log(|\hat{\mathcal{Y}_l}|^{N})$. 

{\bf 2) Restricting the architecture}. The attacker should use a white-box standard architecture, \eg a standard WideResNet~\cite{Zagoruyko2016WRN} in our case study. This constraint reduces the attacker's capability to use the whole information capacity of $\byh$. Because adding any extra components to such standard models could be easily identified. However, it is not straightforward to compute the reduction in the capacity of such restricted $\byh$. 

{\bf 3) Restricting the norm}. The attacker can only observe the softmax output. This constraint means that the sum of all $\byh_n$ must be 1, and thus meaningfully reduces the information capacity of $\byh$. The alphabet size is now reduced from all possible values in a floating-point arithmetic system, \ie $|\hat{\mathcal{Y}_l}|$, to a subset of values that are in the range $[0,1]$: $|\hat{\mathcal{Y}_s}|$. Yet, $\log(|\hat{\mathcal{Y}_s}|^{N-1})$ could be large enough to encode a significant amount of information about $\bx$. 

{\bf 4) Restricting the decimal points}. The attacker can only observe a rounded version of the softmax output. If we round each $\byh_n$ with precision of $q$ decimal digits, then it is possible to precisely enumerate all possible values for the vector $\byh$ according to the {\em stars and bars} pattern in combinatorics~\cite{benjamin2003proofs}. Specifically, the alphabet size $|\hat{\mathcal{Y}_q}|$  of the set of softmax vectors using a finite precision of~$q$ decimal digits is  
\begin{equation}\label{eq:rounded_outputs}
|\hat{\mathcal{Y}_q}| = \binom{10^q+N-1}{N-1}.
\end{equation}
Thus, the exact entropy will be $\log(|\hat{\mathcal{Y}_q}|)$ which can be extremely smaller than $\log(|\hat{\mathcal{Y}_s}|^{N-1})$; but still very large for $q\geq2$ or $N\geq 50$.

{\bf 5) Restricting the entire outputs}. The attacker only observes $\argmax(\byh)$. In this case, no additional information about $\bx$ can leak to the \server; except the \pub task. This setting provides the perfect privacy, but does not allow the \server to perform top-K predictions, to measure the uncertainty in the estimation, to figure out adversarial or out-of-distribution samples, and more~\cite{malinin2018predictive}. 

We summarize the utility-privacy trade-off in an informal but intuitive manner as shown in Table~\ref{tab_ut_pri_trade}. In the end, these defense mechanisms introduce a utility-privacy frontier that can be explored in future studies; for instance to find task-specific noise or rounding parameters that establish desirable trade-offs. However, defending against \vic models raises difficult questions: (i) How to recognize if a \server is \vic or not? (ii) When and to what extent should \users randomize the outputs? (iii) Should we deploy these mechanisms as the default at the \user's side? (iv) How to deal with \servers that continuously collect outputs from each \user?

In general, we believe that there is a need for more investigation on this type of data reconstruction attack, and potential defense mechanisms should be sensible and practical; regarding the envisioned applications in edge computing.

\begin{table}[t]
\centering
\caption{An informal analysis of utility-privacy trade-off for different types of outputs released to the \server.}\label{tab_ut_pri_trade}
\begin{tabular}{@{}lll@{}}
\textbf{Outputs}  & \textbf{Utility} & \textbf{Privacy} \\ \midrule
logits &  Top $N$ Prediction \& Confidence    &  Fragile      \\
softmax & Top $N$ Prediction  &     Weak    \\
rounded & Top $N'\leq N$ Prediction  &  Moderate   \\
argmax  & Top 1 Prediction  &   Perfect.      \\ \bottomrule
\end{tabular}
\end{table}

\subsection{An Optimization-based Defense}

{\bf Motivations.}  Distinguishing \hon models
from \vic ones is not trivial, and blindly applying perturbations to the outputs of all models can damage the utility received from honest servers. Therefore, it would be of great value if we could find a feasible, and ideally low-cost, mechanism to estimate the {\em likelihood of a model being \vic}.  In this context, we present an initial method that can be used by an independent party to assess such a likelihood.

Essentially, the problem of detecting \vic models is kind of a reversed supervised learning problem. For a supervised ML task $\cT$, we train $\cF$ on dataset $(\cX,\cY)$ using an loss function $\mathcal{L}$. Using an ML algorithm, we usually start from a randomly initialized model $\cF_0$ and after some epochs, the model converges to the trained model $\cF$.  However, in our case, we assume that \users observe a trained model $\cF$ that is {\em claimed} to be trained for a task $\cT$ on a dataset $(\cX,\cY)$. But \users do not know what exact objective function $\mathcal{L}$ has been used during training and whether $\cF$ only performs the claimed task $\cT$ or it also secretly performs another task.

\begin{figure}[t]
    \centering
    \includegraphics[width=.7\columnwidth]{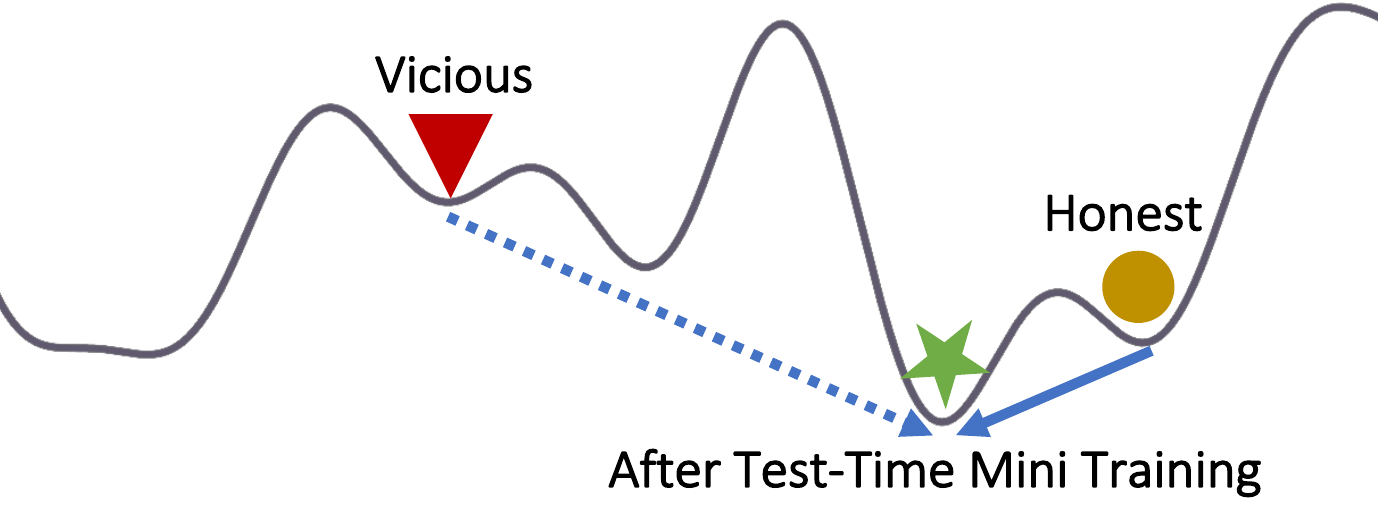}
    \caption{A \vic model, trained to perform a secret task, is expected to deviate more from the optimal solution for the \pub task, compared to an \hon model.}
    \label{fig:test_mini_train}
\end{figure}

{\bf Method.} To estimate the likelihood of a model being \vic, as shown in Figure~\ref{fig:test_mini_train}, we hypothesize that an \hon model has already converged close to the ideal solution for the task $\cT$; if it is only trained using the claimed objective function $\mathcal{L}$. If the model is \vic and is trained to also perform other tasks beyond the claimed one, then it should have not been converged that close to the ideal solution for the task $\cT$. We also assume that the task $\cT$ and the nature of data distribution $(\cX,\cY)$ are public information. To this end, we run mini-batch training on the observed model $\cF$ using a small dataset drawn from $(\cX,\cY)$ and achieve a new trained model $\cF^{+}$. Then, we compare the behavior of the $\cF$ and $\cF^{+}$ to estimate the likelihood of being \vic.

\begin{figure}
\centering
\begin{minipage}[t]{.6\columnwidth}
    \offinterlineskip
    \includegraphics[trim={.0cm 1.0cm .0cm .0cm}, clip, width=\columnwidth]{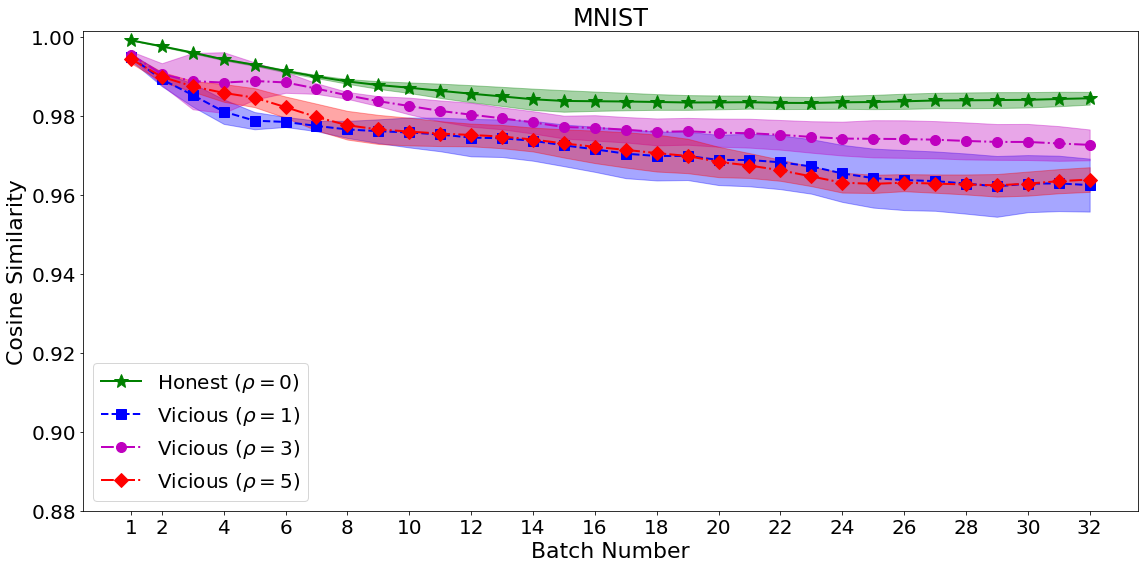}
\end{minipage}
\vfill
\begin{minipage}[t]{.6\columnwidth}
    \offinterlineskip
    \includegraphics[trim={.0cm 1.0cm .0cm .0cm}, clip, width=\columnwidth]{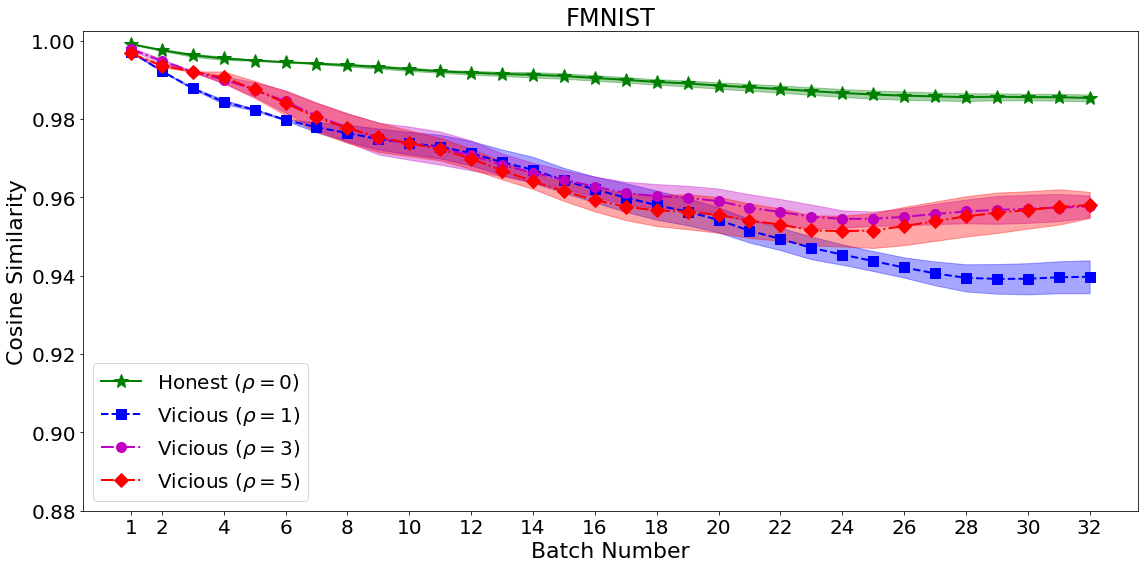}
\end{minipage}
\vfill
\begin{minipage}[t]{.6\columnwidth}
    \offinterlineskip
    \includegraphics[trim={0.0cm 1.0cm .0cm .0cm}, clip, width=\columnwidth]{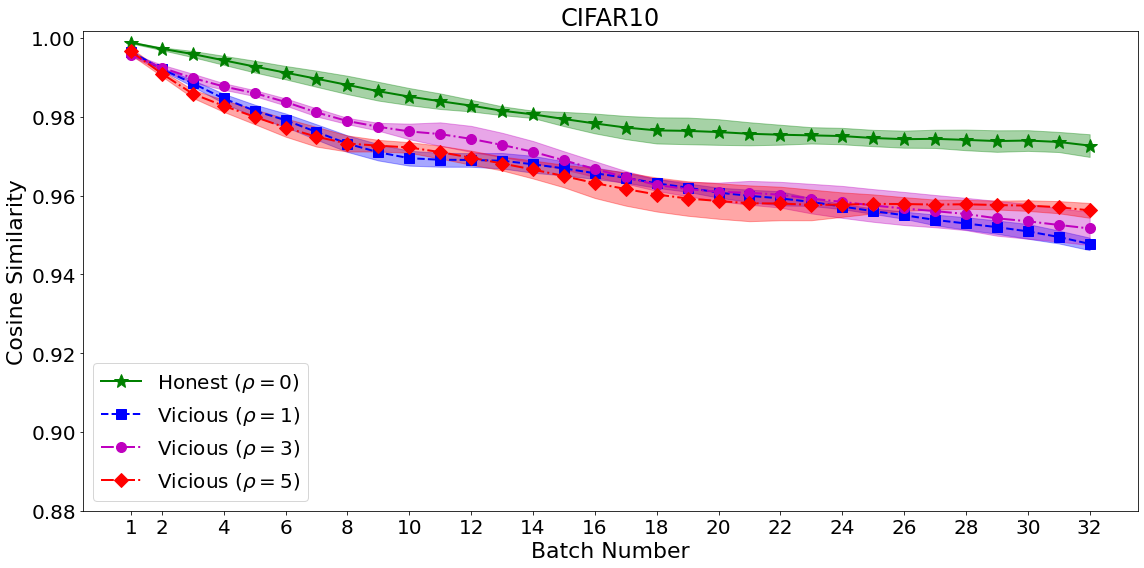}
\end{minipage}
\vfill
\begin{minipage}[t]{.6\columnwidth}
    \offinterlineskip
    \includegraphics[trim={.0cm 1.0cm .0cm .0cm}, clip, width=\columnwidth]{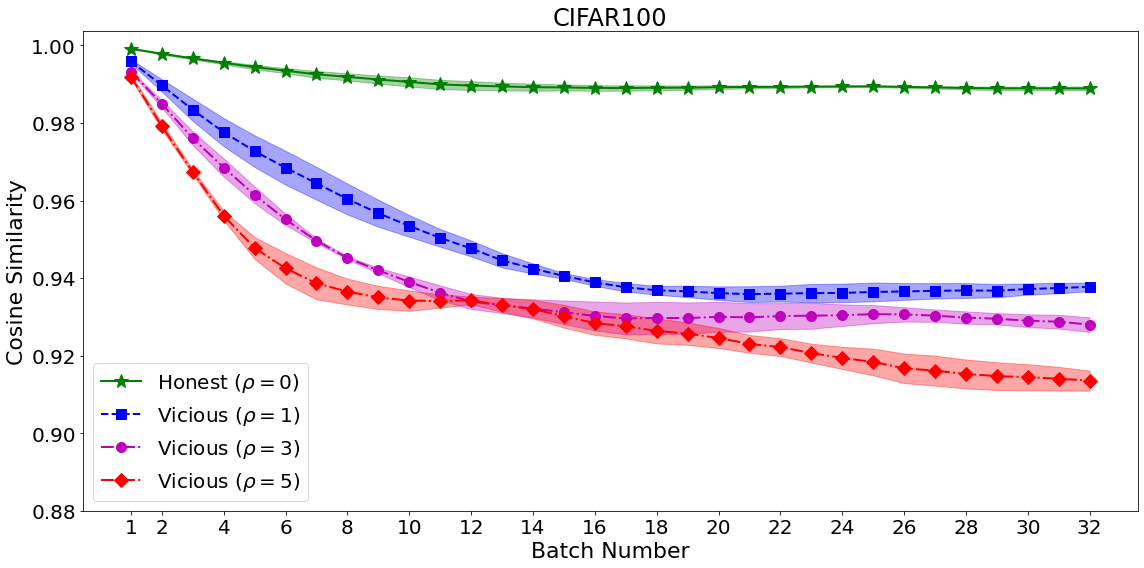}
\end{minipage}
\vfill
\begin{minipage}[t]{.6\columnwidth}
    \offinterlineskip
    \includegraphics[trim={.0cm 0.0cm .0cm .0cm}, clip, width=\columnwidth]{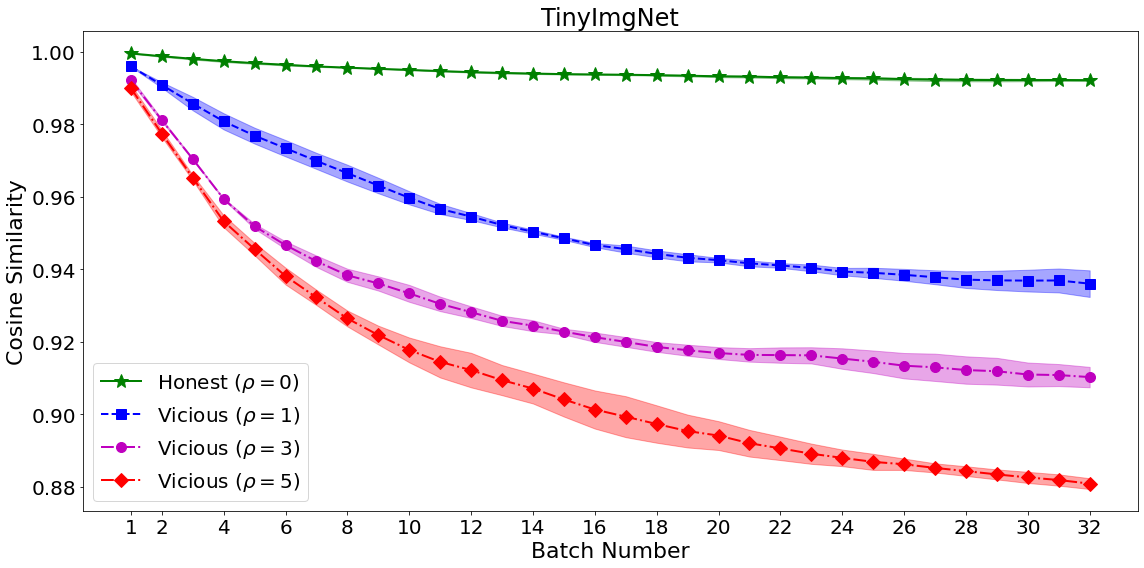}
\end{minipage}
\caption{Cosine similarity between outputs of the received model from the server $\cF$ and the model after mini-batch training $\cF^{+}$. Here, $\rho=\beta^R/\beta^C$ in Table~\ref{tab_results_main}.
}
    \label{fig_model_diff}
\end{figure}

One approach is to compare the difference between $\cF$ and $\cF^{+}$ using some distance measures, like $p$-norm: $\|\cF - \cF^{+} \|_p$. However, it is not necessarily the case that the model's parameters do not change significantly if it is \hon or parameters always change considerably when the model is \vic. For instance, permuting the model's convolutional filters might not change the model's behavior at all, but it affects the distance between the two models. Thus, we use another approach where we compare the outputs of $\cF$ and $\cF^{+}$. In this approach, we know that two \hon models performing the same task $\cT$ can be significantly different in the parameter space but should not show different behavior in their output space.

Let $\bbD^{test}$ be the test dataset (as in \S~\ref{sec_formulation}) and $\bbD^{test+}= \big\{(\bx^k,\by^k)\big\}_{k=1}^{K}$ denote a mini-batch of $K$ new  samples drawn from the data distribution $(\cX,\cY)$. We assume that the \user trains $\cF$ for the \pub task on $\bbD^{test+}$ and obtains $\cF^{+}$. Let  $\byh^{m}=\cF(\bx^m)$ and $\byh^{m+}=\cF^{+}(\bx^m)$ denote the outputs corresponding to the $m$-th sample in $\bbD^{test}$. We compute the {\em cosine similarity}
\begin{equation}\label{eq:cos_sim}
\begin{split}
    \mathcal{C}(\byh, \byh^+) = & \\
    \frac{1}{M}\sum_{m=1}^{M} \Big( & \frac{\byh^{m}\cdot\byh^{m+}} {\|\byh^{m}\|_{2} \|\byh^{m+}\|_{2}}
    =
    \frac{\sum_{n=1}^{N}\byh^{m}_{n}\byh^{m+}_n}{\sqrt{\sum_{n=1}^{N}(\byh^{m}_n)^2} \sqrt{\sum_{n=1}^{N}(\byh^{m+}_n)^2}}\Big),
\end{split}
\end{equation}
and define the likelihood of a model being \vic as
\begin{equation}\label{vic_likelihood}
    \mathsf{v(\cF, \bbD^{test+})} = \frac{1-\mathcal{C}(\byh, \byh^+)}{2} 
\end{equation}

Figure~\ref{fig_model_diff} shows the results of our mini-batch training method for 5 datasets in Table~\ref{tab_results_main}. It is obvious that when the model is \hon, then the cosine similarity $\mathcal{C}(\byh, \byh^+)$ mostly stays very close to $1$; except for CIFAR10, which shows a similarity around 0.98 which is still very high. For datasets with more classes, \eg CIFAR100 and TinyImgNet, after just a few rounds of mini-batch training, a considerable divergence happens; showing that a \vic model can be easily detected. As expected, the more \vic is the model (\ie higher value of $\beta^R/\beta^C$), the smaller the cosine similarity. Overall,  according to Equation~\eqref{vic_likelihood} and our experimental results, a model should be considered as \hon only if $\mathsf{v(\cF, \bbD^{test+})} \leq 0.01$, otherwise it is likely that the model is not trained only for the target task and it might be \vic.

Finally, we emphasize that this defense is an initial effort, serving as a baseline for more practical and realistic defenses. Our objective is to estimate the likelihood of a model being vicious. When it is determined that a model exhibits vicious behavior, the next course of action could be adding some noise or employing other forms of randomization to the logits. 

\end{document}

%% file: _mathdef.tex
\def\cF{\mathcal{F}}
\def\cG{\mathcal{G}}

\def\cS{\mathcal{S}}

\def\cT{\mathcal{T}}
\def\cX{\mathcal{X}}
\def\cY{\mathcal{Y}}

\def\data{\mathbb{D}}
\def\meanD{\mu_{\data}}
\def\MD{\mathsf{d}}
\def\rr{\mathtt{R}}
\def\covM{{\Sigma_{\data}}}

\def\tH{{\mathtt H}}
\def\tI{{\mathtt I}}

\def\bx{{\mathbf x}}

\def\by{{\mathbf y}}

\def\byh{\hat{{\mathbf{y}}}}

\def\bbD{{\mathbb{D}}}

\def\bbR{{\mathbb{R}}}
\def\bbI{{\mathbb{1}}}

\newcommand{\latinphrase}[1]{\textit{#1}}

\newcommand{\ie}{\latinphrase{i.e.,}\xspace}
\newcommand{\eg}{\latinphrase{e.g.,}\xspace}

\newcommand{\aka}{\latinphrase{a.k.a.}\xspace}
\newcommand{\pub}{{target}\xspace}

\newcommand{\priv}{{private}\xspace}

\newcommand{\user}{user\xspace}

\newcommand{\users}{users\xspace}

\newcommand{\Users}{Users\xspace}
\newcommand{\server}{server\xspace}
\newcommand{\servers}{servers\xspace}

\newcommand{\hon}{honest\xspace}

\newcommand{\vic}{vicious\xspace}
\newcommand{\Vic}{Vicious\xspace}

\def\recx{\widetilde{\mathbf x}}

\DeclareMathOperator*{\argmax}{argmax}

\newcommand{\codelink}{
\textcolor{blue}{\mbox{\href{https://github.com/mmalekzadeh/vicious-classifiers}{github.com/mmalekzadeh/vicious-classifiers}}}}